\PassOptionsToPackage{table}{xcolor}
\PassOptionsToPackage{hyphens}{url}
\documentclass[10pt,twocolumn,letterpaper]{article}
\usepackage[pagenumbers]{cvpr}

\def\confName{CVPR}
\def\confYear{2027}

\usepackage{amsmath,amsfonts,bm}

\def\eqref#1{equation~\ref{#1}}
\def\1{\bm{1}}

\DeclareMathAlphabet{\mathsfit}{\encodingdefault}{\sfdefault}{m}{sl}
\SetMathAlphabet{\mathsfit}{bold}{\encodingdefault}{\sfdefault}{bx}{n}

\usepackage[utf8]{inputenc}
\usepackage[T1]{fontenc}
\definecolor{cvprblue}{rgb}{0.21,0.49,0.74}
\usepackage[breaklinks,colorlinks,allcolors=cvprblue]{hyperref}
\usepackage{url}
\usepackage{graphicx}
\usepackage{amsmath,amsfonts,bm}
\usepackage{booktabs}
\usepackage{multirow}
\usepackage{subcaption}
\usepackage[table]{xcolor}
\usepackage{microtype}
\usepackage{tabularx}
\usepackage{array}
\usepackage{caption}
\usepackage{wrapfig}
\usepackage{float}
\usepackage{fontawesome}

\definecolor{colorbest}{RGB}{177, 109, 196}
\definecolor{colorsecond}{RGB}{222, 193, 228}
\definecolor{colorthird}{RGB}{239, 228, 244}
\definecolor{colorrepeat}{HTML}{FEC000}
\definecolor{coloriddeviation}{HTML}{C00200}
\definecolor{colorcopypaste}{HTML}{D86ECC}
\definecolor{cityblue}{RGB}{128, 159, 225}
\newcommand{\first}[0]{\cellcolor{colorbest} }
\newcommand{\second}[0]{\cellcolor{colorsecond}}
\newcommand{\third}[0]{\cellcolor{colorthird}}
\DeclareRobustCommand{\legendsquare}[1]{%
  \textcolor{#1}{\rule{2ex}{2ex}}%
}

\newcommand{\PaperName}{WithEveryone}
\newcommand{\methodname}{\PaperName}

\title{\PaperName: Unified Planning and Identity Grounding for Group Image Generation}

\newcommand{\authorhref}[2]{{\hypersetup{pdfborder={0 0 0}}\href{#1}{#2}}}

\newcommand{\gallerysinbody}{}

\author{
\authorhref{https://doby-xu.github.io/}{\textbf{Hengyuan Xu}}$^{1,2}$
\quad
\authorhref{https://github.com/wangqixun}{\textbf{Qixun Wang}}$^{2,\dag}$
\quad
\textbf{Yiji Cheng}$^{2}$
\quad
\textbf{Miles Yang}$^{2}$
\quad
\textbf{Zhao Zhong}$^{2}$
\\[0.2em]
\textbf{Wei Cheng}$^{3}$
\quad
\textbf{Xingjun Ma}$^{1,\ddag}$
\\[0.4em]
{\normalfont $^{1}$Fudan University \qquad $^{2}$Hunyuan, Tencent \qquad $^{3}$The University of Hong Kong}
\\[0.2em]
{\normalfont\small $^{\dag}$Project lead. \qquad $^{\ddag}$Corresponding author. \qquad
\hypersetup{pdfborder={0 0 0}}\textcolor{cityblue}{%
\faHome~\href{https://doby-xu.github.io/WithEveryone/}{\textbf{Project Page}}
\quad
\faGithub~\href{https://github.com/doby-xu/WithEveryone}{\textbf{Code}}}}
}

\DeclareRobustCommand{\galleryfootnotemark}{}

\makeatletter
\apptocmd{\@maketitle}{\begin{minipage}{\textwidth}
  \centering
  \includegraphics[width=1.0\linewidth]{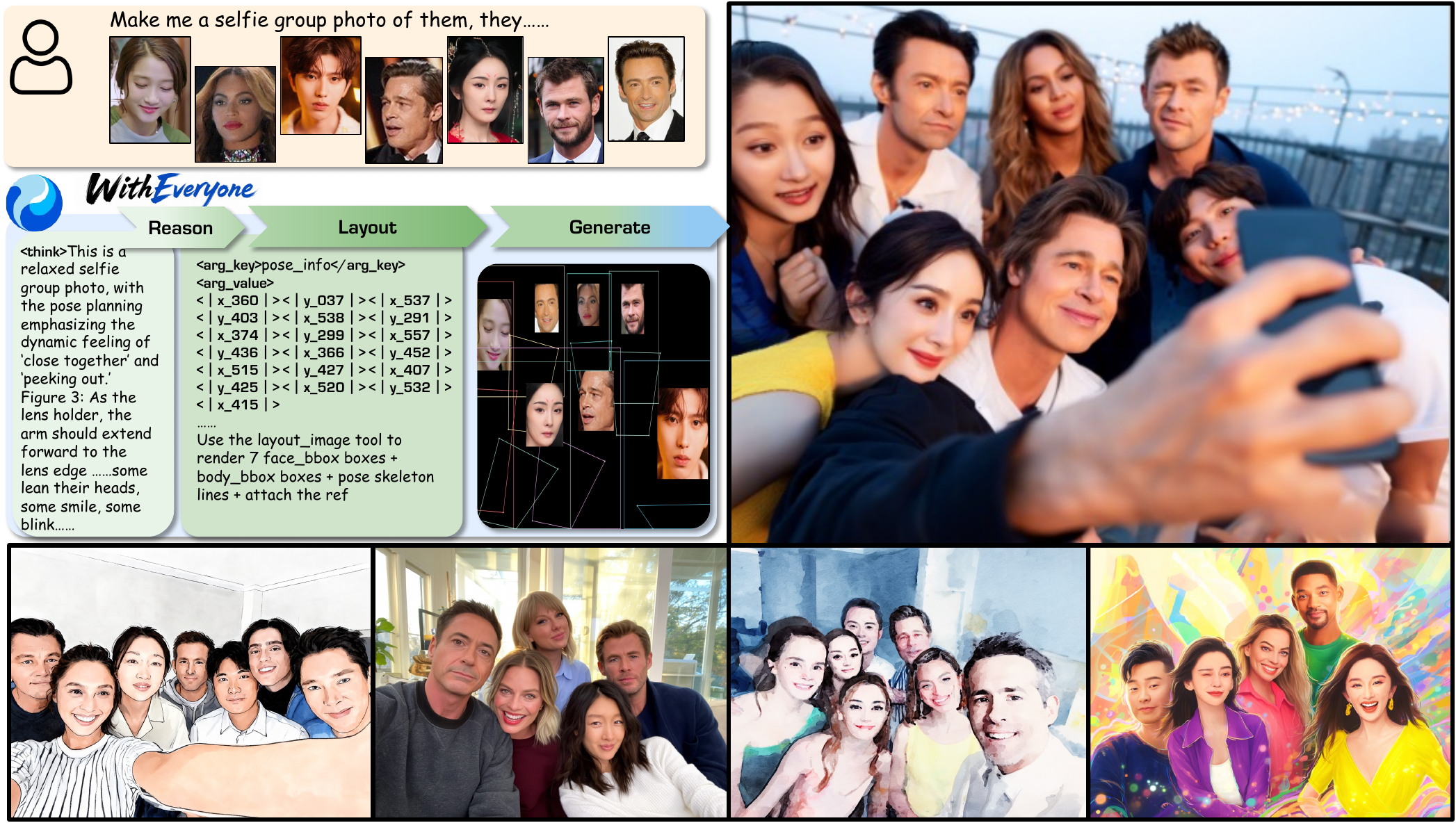}
  \vspace{0.5ex}
  \captionsetup{hypcap=false}
  \captionof{figure}{\small \textbf{Overview of \methodname.}\galleryfootnotemark{} \methodname{} generates coherent group images from five to ten reference identities. A single model reasons about which references participate, plans identity--layout bindings, person regions, and poses, and generates the image conditioned on that plan, so identity and composition are decided in one context rather than by separate modules.}
  \label{fig:teaser}
\end{minipage}
\par\medskip
}{}{\PackageError{paper}{Cannot attach teaser to title}{}}
\makeatother

\begin{document}
\maketitle
\begin{figure*}[p]
  \centering
  \includegraphics[width=0.95\linewidth]{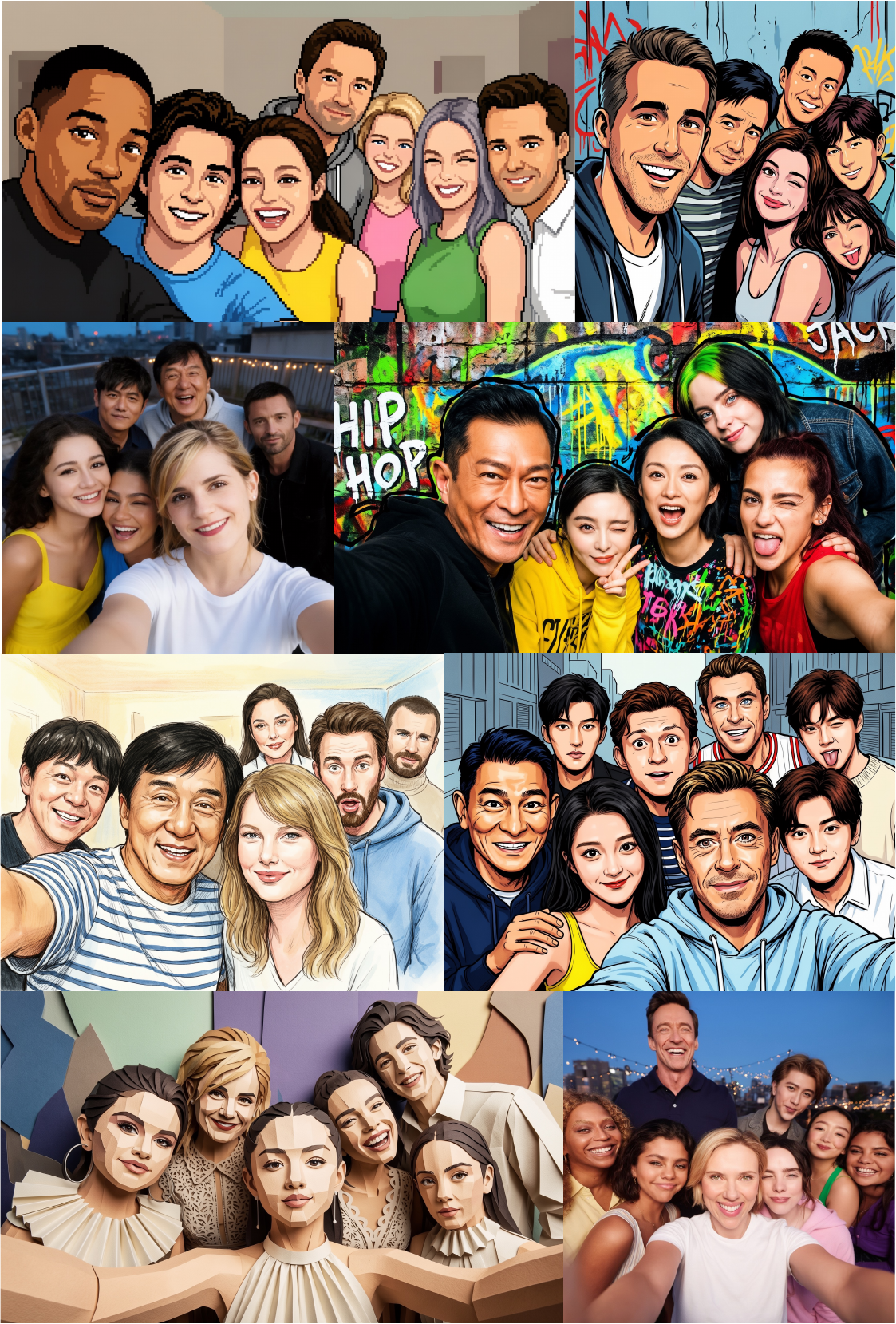}
  \caption{\small \textbf{Group images generated by \methodname.} The samples cover different group sizes, scene types, and visual styles.}
  \label{fig:gallery1}
\end{figure*}

\begin{abstract}
    Identity-preserving image generation becomes increasingly unreliable when a scene must contain many specified people. Beyond retaining each identity, the model must bind every reference to a distinct person and location, while training-time identity losses must establish correspondence among several noisy predicted faces. We introduce \methodname{}, a unified framework for generating group images of 5--10 reference identities. \methodname{} injects each selected identity as an addressed token, predicts a structured identity--layout plan, and renders the plan as a visual condition. Its key objective, Layout-Grounded ID Loss, uses annotated face regions to supervise the intended identities directly, avoiding unstable embedding-based face matching; ID Representation Forcing additionally trains a prediction for each identity before image synthesis. \methodname{} leads in identity similarity (0.538 versus GPT-Image-2.5's 0.472), with lower Copy-Paste (0.149 versus 0.197), 94.5\% identity coverage, and 9.6\% identity collisions. These results show that explicit identity--layout grounding enables identity-preserving generation to scale to larger groups without relying on direct reference-face copying.
    \end{abstract}
\section{Introduction}

\begin{figure*}[t]
  \centering
  \includegraphics[width=\textwidth]{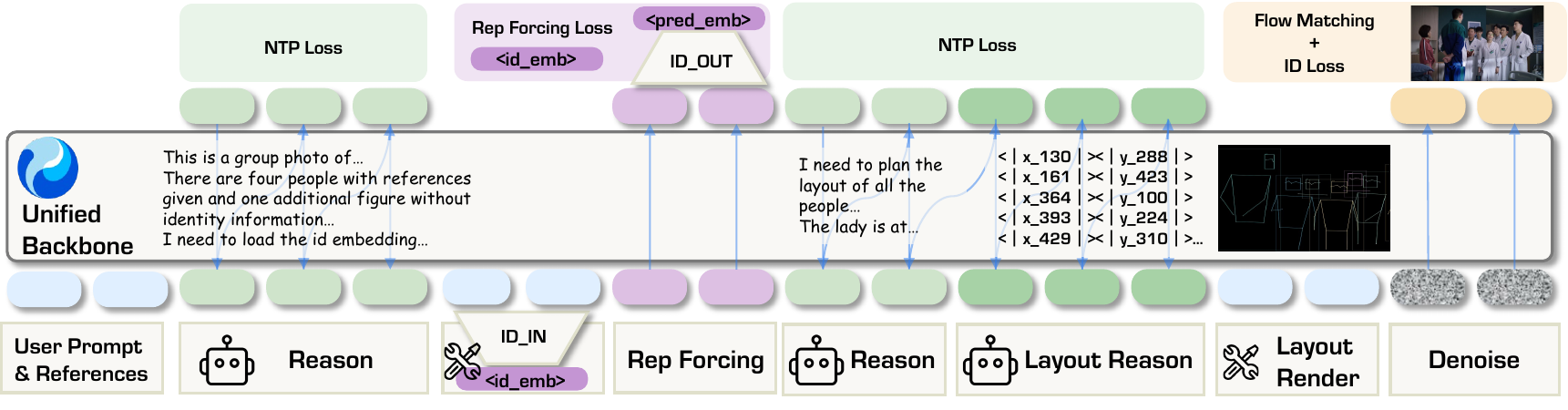}
  \caption{Overview of \methodname. The model loads selected reference identities as ID tokens, predicts structured Layout CoT, renders a visual layout condition, and predicts identity representations before generating the target group image. Modality-boundary tokens such as ``BOI'' are omitted, and the actual CoT order follows the text.}
  \label{fig:pipeline}
\end{figure*}

Recent image generation models, especially unified multimodal models (UMMs), have advanced rapidly in visual fidelity, text understanding, and controllability, and identity-preserving generation further lets users specify a person through one or more reference images \citep{li2024photomaker, wang2024instantid}. However, existing open academic methods and benchmarks primarily address one or a small number of people, typically covering one to five reference identities (IDs) \citep{xu2025withanyone, cheng2025umo, borse2025multihuman}; even strong proprietary systems such as Nano Banana Pro support at most five people \citep{raisinghani2025nanobananapro}. Real group-image creation requires jointly generating more specified people rather than repeatedly generating independent single-person images. We study group-image generation from five to ten reference IDs in one semantically and visually coherent image.

As the number of reference IDs increases, identity similarity is widely reported to decrease \citep{xu2025withanyone,kim2024instantfamily,zhang2025idpatch,he2024uniportrait}, and models tend to duplicate faces or merge IDs in multi-person scenes \citep{borse2026resolving}. When more reference images, visual tokens, and identity conditions share one generation context, each identity signal is more easily diluted, and cross-identity interference and incorrect binding become more likely. Supplying reference images or identity embeddings as conditions therefore does not guarantee that the model keeps reading every identity during generation: large-group generation needs a mechanism that both provides each reference identity and constrains the context to retain its representation.

Prior work also shows that conditioning alone is not enough, and that supervising the generated face directly with an identity loss is necessary for strong identity preservation \citep{guo2024pulid,xu2025withanyone}. Such losses, however, are formulated for a single person or at most two. Extending them to a large group is not a matter of running the same loss more times, because the loss must first decide which generated face belongs to which reference, and existing formulations recover that correspondence by matching face embeddings in the generated image. Early in training or at high noise levels, however, the predicted faces are nearly identical, so this matching becomes unreliable and the identity loss fails or even becomes counterproductive.

More references also complicate spatial and bodily relationships. Identity similarity answers who is generated, but not where people appear, what poses they take, or how they are organized. As the group grows, pairwise relations multiply and one misplaced person propagates errors through the surrounding composition, making overlaps, limb errors, and crowded layouts especially likely. We therefore offload this planning burden to the understanding side of a unified model before image generation.

To address these challenges, we propose \methodname, a UMM framework shown in Figure~\ref{fig:teaser}. On the identity side, we extract an identity embedding for each reference person, project it into an ID token, and bind that token to the planned person it should control, so that references form an addressed set rather than an unordered pool. Because ID tokens are easily ignored in long multi-ID sequences, \emph{ID Representation Forcing} adapts representation forcing \citep{wang2026representation} to identities, requiring the model to predict a representation aligned with each target person before synthesis. At the output level, we introduce a flow-compatible \emph{Layout-Grounded ID Loss} (LG-ID Loss), which removes the matching problem described above rather than approximating it: the correspondence is read off the layout annotation, so each identity is supervised in its own annotated region no matter how many people share the image.

On the understanding side, the model autoregressively predicts a structured \emph{Layout Chain of Thought} (Layout CoT) containing person indices, bounding boxes, identity--layout bindings, and pose keypoints, extending ATLAS-style planning \citep{liu2026atlas} to an identity-aware setting in which every planned person is tied to a reference identity; a deterministic renderer then converts the plan into a visual condition. ID tokens and Representation Forcing determine who to generate, Layout CoT determines where and how each person appears, and identity--layout binding joins the two in a single plan.

We evaluate \methodname{} against academic, open-source, and proprietary systems, including GPT-Image-2 and GPT-Image-2.5. Ours reaches the highest Sim(Tgt)/Sim(Ref), at 0.538/0.641, compared with 0.472/0.615 for GPT-Image-2.5, while lowering Copy-Paste~\citep{xu2025withanyone} from 0.197 to 0.149. This identity advantage comes with competitive scene similarity, but does not extend to every quality or coverage metric. Our contributions are threefold:
\begin{itemize}
  \item We propose interleaved identity and layout reasoning inside a single unified model: each selected reference is loaded as an ID token, a structured Layout CoT extends ATLAS-style planning with identity--layout binding and joint prediction of face regions, body extents, and poses, and a deterministic renderer turns the plan into a visual condition. ID Representation Forcing adapts representation forcing by adding supervised per-identity predictions before image synthesis, encouraging the selected identities to remain addressable in the shared context.
  \item We propose the LG-ID Loss, which grounds output-side identity supervision in the layout annotation instead of in embedding-based face matching. This removes the failure mode that prevents existing identity losses from being applied beyond two people, and it is the single largest source of identity gain in our system.
  \item We build and systematically evaluate a group-image generation system for five to ten people. On an identity-disjoint benchmark it achieves higher target-context identity similarity than GPT-Image-2.5 (Sim(Tgt) 0.538 versus 0.472) at a lower copy-paste artifact score (0.149 versus 0.197), and our analysis separates plan prediction from plan execution as distinct sources of the remaining error.
\end{itemize}

\section{Related Work}
\label{sec:related}

\paragraph{Identity-Preserving and Identity-Consistent Generation.}
Identity-preserving generation has progressed from single-person methods~\citep{wang2024instantid,li2024photomaker,valevski2023face0,wang2024stableidentity,yan2023facestudio,xiao2025fastcomposer} to multi-person customization~\citep{zhang2025idpatch,he2024uniportrait,kim2024instantfamily,cheng2025umo}. Existing methods, however, remain limited to small groups and can trade identity similarity for copy-paste behavior~\citep{chen2025xverse,mou2025dreamo,jiang2025infiniteyou,xu2025withanyone}; reported examples from WithAnyone and UMO include at most four and three people, respectively. Output-side identity supervision is similarly limited: PuLID~\citep{guo2024pulid} handles one identity, while WithAnyone uses embedding-based Hungarian matching for at most two. Under training-time noise, this matching becomes unreliable as the number of faces grows. Our LG-ID Loss instead obtains correspondence from layout annotations, enabling output-side supervision for five to ten people.

%A second line of work supervises intermediate context instead of outputs: representation forcing~\citep{wang2026representation} aligns an intermediate prediction so that the shared context carries information the generator will need, and we specialize it to identities by attaching one supervised prediction per reference person rather than a single global representation.

\paragraph{Text-Image-to-Image Models and Unified Models.}
Reference-conditioned generation is now supported by open models~\citep{batifol2025flux,wu2025qwenimage,team2025longcat,diao2026sensenova,cai2026hidream} and proprietary systems~\citep{raisinghani2026nanobanana2,raisinghani2025nanobananapro,bytedanceseed2026seedream5pro,openai2025gpt4o,openai2026gptimage2}. Unified architectures such as HunyuanImage~3 and BAGEL generate text and images in a shared context, enabling in-context planning before synthesis~\citep{cao2025hunyuanimage,deng2025bagel,zhou2025transfusion}.

\paragraph{Layout Planning and Generation.}
Most layout-conditioned methods use a separate planner and image generator~\citep{feng2023layoutgpt,gupta2023visual,yang2024mastering,zhang2024realcompo,fang2025got,duan2025got,borse2026ar2can}, whereas ATLAS and PlanGen unify planning and synthesis in one contextual sequence~\citep{liu2026atlas,he2025plangen}. We extend this setting from text-to-image to identity-conditioned generation: each planned person is bound to a reference identity, with jointly predicted face regions, body extents, and poses rendered as a visual condition.
\section{Method}
\label{sec:method}

\subsection{Overview}

We consider group-image generation from a text prompt and a set of five to ten reference identities. Our model follows a transfusion-style \citep{zhou2025transfusion} mixture-of-transformers multimodal architecture \citep{deng2025bagel,cao2025hunyuanimage} in which text and structured reasoning are predicted autoregressively, while target-image latents are learned through flow matching \citep{lipman2022flow}. This shared causal context lets high-level decisions made before image generation directly condition the subsequent image tokens.

As shown in Figure~\ref{fig:pipeline}, \methodname{} prepares three complementary conditions before synthesizing the target image: it selects the reference identities participating in the scene and loads a compact representation for each, predicts a structured multi-person layout together with the association between identities and planned people, and finally predicts target identity representations from the accumulated context. The rendered layout condition and the representation scaffold jointly condition flow-based image generation.

\subsection{Interleaved Identity and Layout Reasoning}
\label{sec:cot}

\paragraph{ID binding and Representation Forcing.}
The model selects requested reference people while allowing unreferenced people, and complements low-level image features with a high-level identity signal, analogous to ViT/VAE features~\citep{dosovitskiy2020image}: each selected identity's 512-dimensional ArcFace embedding~\citep{deng2019arcface,Deng2020retinaface} is projected through an MLP into an ID token and bound in the subsequent layout. Because these sparse tokens can be diluted in a 10--20K-token context, ID Representation Forcing~\citep{wang2026representation} places one prediction token per target-linked identity before image synthesis and maps its contextual hidden state $\mathbf{h}^{\mathrm{rep}}_i$ back to the ArcFace space:
\begin{equation}
  \hat{\mathbf{e}}_i = g_{\mathrm{out}}(\mathbf{h}^{\mathrm{rep}}_i), \qquad
  \mathcal{L}_{\mathrm{RF}} =
  \frac{1}{M}\sum_{i=1}^{M}\left(1-\cos\left(\hat{\mathbf{e}}_i,\mathbf{e}^{\mathrm{tgt}}_i\right)\right),
\end{equation}
where $g_{\mathrm{out}}$ is an MLP head, $\mathbf{e}^{\mathrm{tgt}}_i$ is the target-person embedding, and $M$ is the number of supervised identities. Unlike next-token prediction or flow matching, this loss directly aligns a continuous identity prediction from the shared context; the resulting scaffold stays causally available to later image tokens (Appendix~\ref{sec:appendix-id-details}).

\paragraph{Structured Layout CoT.}
ID-preserving generation must also resolve the spatial and bodily relationships among people, which we formulate as a structured Layout CoT. Following ATLAS~\citep{liu2026atlas}, the model autoregressively predicts identity--layout bindings, person and face regions, body extents, and pose keypoints using a discretized coordinate vocabulary with 2,002 tokens, corresponding to 1,001 positions on each axis. The stages are emitted in a fixed causal order, so each decision is conditioned on the identities and regions already committed; short natural-language connectors carry the scene state across stages, while the spatial fields remain parseable and directly supervised. The model then calls a rendering tool that draws the planned face regions, body regions, and pose skeletons onto a blank canvas at the target aspect ratio, which is inserted into the context as a condition image.

\subsection{Training Data and Objectives}
\label{sec:training}

\begin{figure*}[t]
  \centering
  \includegraphics[width=\textwidth]{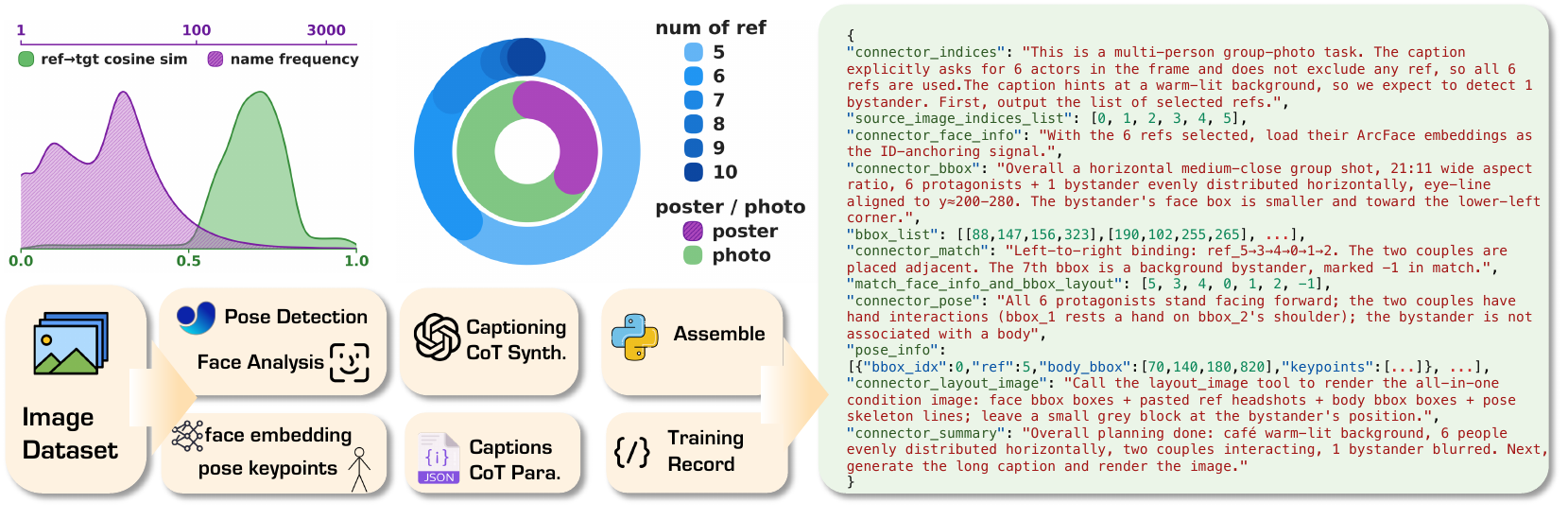}
  \caption{\textbf{Data statistics, pipeline, and example.} We report the distributions of identity occurrence frequency, reference count, and real photographs versus collage and poster images; the data-construction pipeline; and an example training sample. The corpus contains 400K group-image samples, and the reference-count distribution covers the five-to-ten-person range used at evaluation time.}
  \label{fig:data-pipeline}
\end{figure*}

\paragraph{Training data and sequence.}
Each training sample interleaves reference images and the prompt, identity selection and ID loading, layout planning, the rendered layout condition, a recaption, the identity prediction, and the target image, in that causal order; Figure~\ref{fig:data-pipeline} shows how these signals are derived from real group photographs, and Appendix~\ref{sec:appendix-data} gives the full sequence and annotation pipeline.

\paragraph{Layout-Grounded ID Loss.}
Existing output-side identity losses~\citep{guo2024pulid,xu2025withanyone} either handle one person or recover correspondence by matching generated-face embeddings, which becomes unstable when noisy one-step predictions contain five to ten barely distinguishable faces. PuLID avoids correspondence by supervising one identity, whereas WithAnyone uses Hungarian matching and reports it for at most two. At high noise, arbitrary pairings pull one identity toward another, cancelling rather than sharpening supervision. We remove the matching problem rather than improving the matcher: we take each identity's annotated face box from the target image and crop the predicted image at the same box, so every crop pair is correct by construction for ArcFace supervision~\citep{deng2019arcface}. This holds only at sufficiently low noise, where generated faces are recognizable and follow their boxes; a timestep threshold enforces this (Appendix~\ref{sec:appendix-idloss}), and plan adherence converges before identity preservation (Section~\ref{sec:experiment-analysis}). Because correspondence no longer depends on generated appearance, \textbf{Layout-Grounded ID Loss} (LG-ID Loss) scales independently of group size and provides the largest individual identity gain in our ablation (Section~\ref{sec:ablations}).

\paragraph{Joint objective.}
Different parts of the sequence receive supervision appropriate to their representation. Connector text, structured layout tokens, summaries, and recaptions use next-token prediction; target-image latents use flow matching; and identity predictions use the cosine alignment above. Given a noisy latent $\mathbf{x}_t$ and predicted velocity $\mathbf{v}_{\theta}(\mathbf{x}_t,t)$ under the reverse-flow convention, the one-step clean-image estimate used by the LG-ID Loss is
\begin{equation}
  \hat{\mathbf{x}}_{\mathrm{clean}} = \mathbf{x}_t - t\,\mathbf{v}_{\theta}(\mathbf{x}_t,t).
\end{equation}
The overall objective is
\begin{equation}
  \mathcal{L} =
  \mathcal{L}_{\mathrm{NTP}} +
  \lambda_{\mathrm{FM}}\mathcal{L}_{\mathrm{FM}} +
  \lambda_{\mathrm{RF}}\mathcal{L}_{\mathrm{RF}} +
  \lambda_{\mathrm{ID}}\mathcal{L}_{\mathrm{ID}}.
\end{equation}
We set $\lambda_{\mathrm{FM}}=1.0$, $\lambda_{\mathrm{RF}}=1.0$, and $\lambda_{\mathrm{ID}}=0.5$; Appendix~\ref{sec:appendix-training-config} lists the backbone, initialization, data scale, optimizer, and hardware. At inference time, the model predicts identity selection, Layout CoT, and recaption autoregressively; the renderer deterministically constructs the layout condition from the predicted plan. Target identity embeddings are used only as training targets for representation forcing, not as inference-time conditions.

\section{Experiments}
\label{sec:experiments}

\subsection{Experimental Setup}
\label{sec:experimental-setup}

\paragraph{Benchmark.}
We evaluate on 210 real group images with five to ten reference identities, stratified into 60, 50, 40, 30, 20, and 10 examples, respectively. Each target contains exactly one detected face per reference and no duplicate identities. All benchmark identities and any training images containing them are held out, so the comparison tests unseen people rather than image-level memorization.

\paragraph{Baselines and protocol.}
We compare academic identity methods (UMO~\citep{cheng2025umo}, WithAnyone~\citep{xu2025withanyone}, UniPortrait~\citep{he2024uniportrait}, DreamO~\citep{mou2025dreamo}, and ID-Patch~\citep{zhang2025idpatch}), open models (FLUX.2 Klein~\citep{flux-2-2025}, Qwen-Image-Edit~\citep{wu2025qwenimage}, LongCat-Image-Edit~\citep{team2025longcat}, OmniGen2~\citep{wu2025omnigen2}, SenseNova U1~\citep{diao2026sensenova}, BAGEL~\citep{deng2025bagel}, and HiDream-O1~\citep{cai2026hidream}), and proprietary Nano Banana Pro~\citep{raisinghani2025nanobananapro} and Nano Banana 2~\citep{raisinghani2026nanobanana2}, Seedream~\citep{bytedanceseed2026seedream5pro}, and GPT-Image-2~\citep{openai2026gptimage2} and GPT-Image-2.5 (Sunburst)~\citep{openai2026gptimage25}. Ours is evaluated at 2K in Table~\ref{tab:main-results}. We exclude missing generations (e.g., API-censored outputs) and undefined scores. Ablation experiments use earlier checkpoints without post-training and are evaluated at 1K. ID-Patch and WithAnyone receive ground-truth layouts because they do not plan them.

\paragraph{Metrics.}
Following WithAnyone~\citep{xu2025withanyone}, $\mathrm{Sim(Ref)}$ and $\mathrm{Sim(Tgt)}$ average ArcFace (buffalo\_l)~\citep{deng2019arcface}, FaceNet~\citep{schroff2015facenet}, and AdaFace~\citep{kim2022adaface}; ablations use ArcFace only. Assigning generated faces to references is reliable on the clean final image; the LG-ID Loss instead addresses noisy one-step training predictions. Human-validated Copy-Paste~\citep{xu2025withanyone} measures reuse; Coverage is the mean per-image fraction of valid references with a one-to-one Hungarian match of at least 0.20; and Dup is the fraction whose best generated face is also selected by another reference. DINO-I~\citep{oquab2023dinov2} and CLIP-I/T~\citep{radford2021learning} measure image/text alignment; Aes uses aesthetic-predictor-v2.5~\citep{aesthetic-predictor-v2-5}, and HPSv3~\citep{ma2025hpsv3widespectrumhumanpreference} reports the unnormalized preference score with English benchmark captions; higher is better for both. Count is count accuracy; and Plan IoU is mean matched IoU between supplied face boxes and detected generated faces, measuring own-plan or oracle-plan execution. Appendix~\ref{sec:appendix-experimental-details} gives full definitions.

\begin{table*}[t]
  \centering
  \caption{\small \textbf{Quantitative comparison on our 5--10-person benchmark.} The benchmark contains 210 examples; means use available valid scores and metric-specific sample counts. Identity similarities average the three face encoders of the main protocol. Coverage is the fraction of references whose one-to-one matched generated face reaches a similarity of 0.20, and Dup the fraction that collapse onto a face already claimed by another identity. \legendsquare{colorbest}, \legendsquare{colorsecond}, and \legendsquare{colorthird} mark the first-, second-, and third-best results; for Copy-Paste only the three methods with the highest Sim(Ref) are ranked, since lower similarity means less copying naturally. WithAnyone and ID-Patch use ground-truth layouts and are excluded from the Coverage and Dup rankings.}
  \label{tab:main-results}
  \small
  \setlength{\tabcolsep}{3.5pt}
  \resizebox{\textwidth}{!}{%
  \begin{tabular}{l|ccc|cc|ccccc}
    \toprule
    \multirow{2}{*}{\textbf{Method}} &
    \multicolumn{3}{c|}{\textbf{Identity Similarity}} &
    \multicolumn{2}{c|}{\textbf{Identity Coverage}} &
    \multicolumn{5}{c}{\textbf{Generation Quality}} \\
    \cmidrule(r){2-4}\cmidrule(lr){5-6}\cmidrule(l){7-11}
    & Sim(Tgt) $\uparrow$ & Sim(Ref) $\uparrow$ & Copy-Paste $\downarrow$
    & Coverage $\uparrow$ & Dup $\downarrow$
    & CLIP-I $\uparrow$ & DINO-I $\uparrow$ & CLIP-T $\uparrow$ & Aes $\uparrow$ & HPSv3 $\uparrow$ \\
    \midrule
    \multicolumn{11}{l}{\emph{Academic identity-preserving methods}} \\
    WithAnyone & 0.405 & 0.483 & 0.096 & 0.957 & 0.045 & 0.807 & 0.695 & 0.281 & 5.34 & 11.241 \\
    UMO & 0.371 & 0.484 & 0.112 & 0.630 & 0.258 & 0.780 & 0.663 & 0.286 & 5.03 & 10.747 \\
    UniPortrait & 0.339 & 0.464 & 0.115 & 0.635 & 0.187 & 0.679 & 0.415 & \second{0.301} & 5.09 & 10.286 \\
    DreamO & 0.297 & 0.331 & 0.027 & 0.298 & 0.299 & 0.724 & 0.611 & 0.282 & 5.31 & 10.299 \\
    ID-Patch & 0.225 & 0.259 & 0.032 & 0.247 & 0.224 & 0.609 & 0.320 & \first{0.331} & 4.89 & 6.446 \\
    \midrule
    \multicolumn{11}{l}{\emph{Open-source general-purpose models}} \\
    FLUX.2 Klein & 0.264 & 0.265 & 0.002 & 0.314 & 0.265 & 0.787 & 0.685 & \third{0.291} & 5.30 & \first{14.149} \\
    Qwen-Image-Edit & 0.253 & 0.289 & 0.018 & 0.176 & 0.291 & 0.614 & 0.312 & 0.227 & \first{5.96} & 8.081 \\
    LongCat-Image-Edit & 0.287 & 0.288 & $-$0.001 & 0.381 & 0.264 & 0.802 & 0.657 & 0.288 & 5.31 & 11.401 \\
    OmniGen2 & 0.267 & 0.275 & $-$0.002 & 0.418 & 0.267 & 0.774 & 0.633 & 0.288 & 5.43 & 11.886 \\
    SenseNova U1 & 0.253 & 0.243 & $-$0.009 & 0.213 & 0.256 & 0.797 & 0.686 & 0.289 & 5.25 & 11.445 \\
    BAGEL & 0.223 & 0.225 & 0.003 & 0.136 & 0.276 & 0.752 & 0.635 & 0.282 & 4.10 & 10.358 \\
    HiDream-O1 & 0.353 & 0.376 & 0.026 & 0.780 & 0.190 & 0.806 & 0.692 & 0.286 & \third{5.49} & 12.004 \\
    \midrule
    \multicolumn{11}{l}{\emph{Proprietary systems}} \\
    Nano Banana Pro & 0.453 & 0.478 & 0.041 & 0.674 & 0.148 & 0.839 & 0.712 & 0.275 & 5.39 & 12.192 \\
    Nano Banana 2 & 0.451 & 0.480 & 0.045 & 0.884 & 0.099 & \first{0.860} & \first{0.731} & 0.276 & 5.28 & \third{12.280} \\
    GPT-Image-2 & \third{0.462} & \third{0.583} & \second{0.169} & 0.905 & \second{0.075} & \second{0.853} & \third{0.719} & 0.270 & 5.36 & 12.150 \\
    GPT-Image-2.5 & \second{0.472} & \second{0.615} & \third{0.197} & \second{0.925} & 0.106 & 0.839 & 0.710 & 0.279 & \second{5.63} & 12.023 \\
    Seedream 4.5 & 0.407 & 0.495 & 0.114 & 0.859 & 0.175 & 0.829 & 0.698 & 0.286 & 5.29 & 11.849 \\
    Seedream 5.0 Pro & 0.436 & 0.522 & 0.114 & \third{0.913} & \first{0.065} & \third{0.850} & 0.715 & 0.275 & 5.44 & \second{12.638} \\
    \midrule
    \methodname{} & \first{0.538} & \first{0.641} & \first{0.149} & \first{0.945} & \third{0.096} & 0.849 & \second{0.724} & 0.275 & 5.21 & 11.915 \\
    \bottomrule
  \end{tabular}}
\end{table*}

\subsection{Cross-Model Comparison}
\label{sec:main-results}

\paragraph{Identity, coverage, and copying.}
Table~\ref{tab:main-results} shows that \methodname{} combines the strongest target-context and reference identity similarity with less copying than either GPT-Image baseline. This combination preserves identity while adapting faces to the target scene. It also achieves the highest coverage among methods without ground-truth layouts, showing that the identity gain extends to composing the requested group.

\begin{figure}[t]
  \centering
  \includegraphics[width=\linewidth]{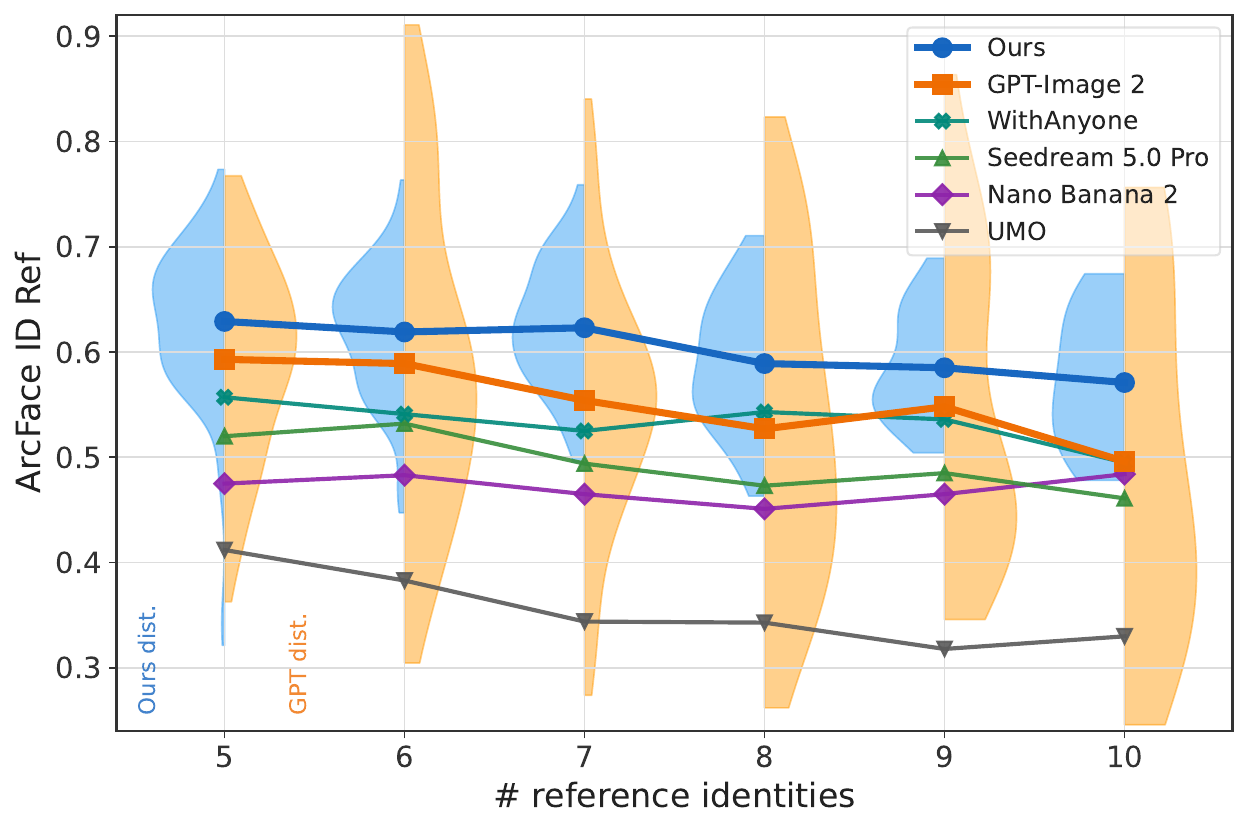}
  \caption{\small \textbf{Scaling to more references.} Lines show mean ArcFace similarity to the references; distributions are shown for \methodname{} and GPT-Image-2.}
  \label{fig:scaling}
\end{figure}

\paragraph{Generation quality and scaling.}
The identity gains accompany competitive scene similarity, including the second-highest DINO-I. Although its Aes and HPSv3 scores are lower than those of the GPT-Image models, \methodname{} remains competitive on both metrics. Across five to ten references (Figure~\ref{fig:scaling}), \methodname{} ranks first at every group size and declines from 0.629 to 0.571, versus 0.593 to 0.496 for GPT-Image-2 and 0.412 to 0.330 for UMO; its distribution is tighter, while GPT-Image-2 has heavier tails. Together, the group-size trends show sustained identity preservation as more references compete within the same scene (Appendix~\ref{sec:appendix-main-reading}).

\begin{table}[t]
  \centering
  \caption{\small \textbf{Ablation results.} The rows form three incremental branches from the same base: layout (P1--P3), ID token and Representation Forcing (P1, P4, P5), and LG-ID Loss (P1, P6); P7 is the complete configuration and additionally uses the text-to-layout corpus.}
  \label{tab:ablations}
  \small
  \setlength{\tabcolsep}{3pt}
  \resizebox{\linewidth}{!}{%
  \begin{tabular}{llcccc}
    \toprule
    & Variant & Sim(Ref) $\uparrow$ & Sim(Tgt) $\uparrow$ & Count $\uparrow$ & Coverage $\uparrow$ \\
    \midrule
    P1 & Default & 0.339 & 0.304 & 0.771 & 0.741 \\
    P2 & Model Layout & 0.364 & 0.316 & 0.828 & 0.813 \\
    P3 & GT Layout & 0.412 & 0.367 & \textbf{0.958} & 0.891 \\
    P4 & ID Token & 0.351 & 0.313 & 0.827 & 0.761 \\
    P5 & ID Token + RF & 0.364 & 0.328 & 0.817 & 0.782 \\
    P6 & LG-ID Loss only & 0.506 & 0.435 & 0.845 & 0.947 \\
    \rowcolor{gray!12}
    P7 & Full Model & \textbf{0.555} & \textbf{0.461} & 0.869 & \textbf{0.960} \\
    \bottomrule
  \end{tabular}}
\end{table}

\begin{figure*}[t]
  \centering
  \includegraphics[width=0.95\textwidth]{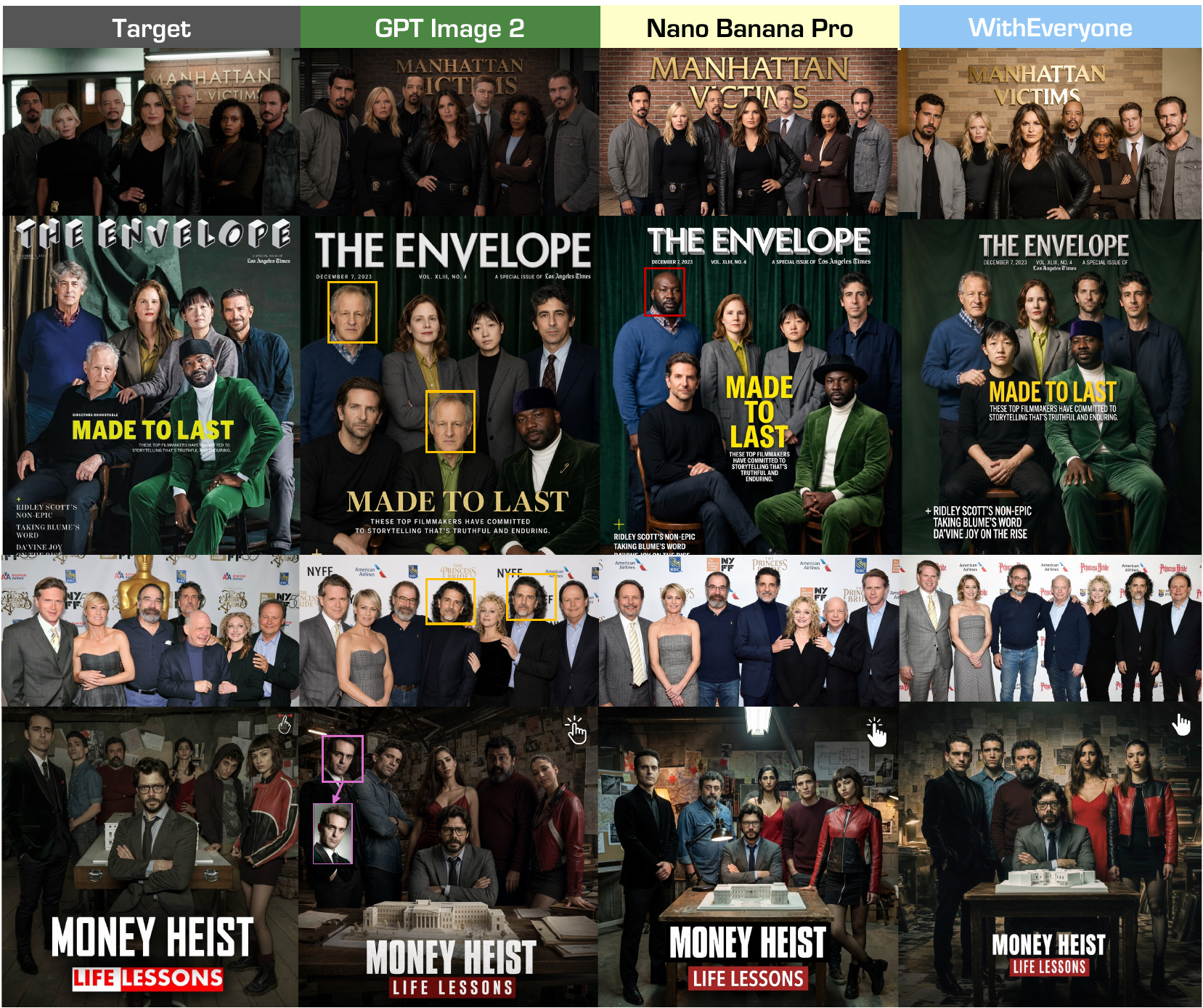}
  \caption{\textbf{Qualitative comparison with proprietary models.} \legendsquare{colorrepeat}, \legendsquare{coloriddeviation}, and \legendsquare{colorcopypaste} mark faces that reproduce a person already generated elsewhere in the image, that are not recognizable as their bound reference, and that appear transplanted without adapting to the pose, lighting, or viewpoint of the scene. These author annotations are illustrative; their quantitative counterparts are the duplicate rate, Sim(Ref), and Copy-Paste.}
  \label{fig:quality}
\end{figure*}

\subsection{Ablation Studies}
\label{sec:ablations}

Table~\ref{tab:ablations} evaluates layout conditioning (P1--P3), ID tokens and Representation Forcing (P1, P4, P5), LG-ID Loss (P1, P6), and the full model (P7). Variants are trained independently under the same recipe, except that P7 also uses the text-to-layout corpus (Appendix~\ref{sec:appendix-data}); Appendix~\ref{sec:appendix-ablation-details} gives the full setup of each branch.

\paragraph{Effect of the LG-ID Loss.}
LG-ID Loss provides the largest individual gain, raising Sim(Ref)/Sim(Tgt) from 0.339/0.304 to 0.506/0.435 and Count/Coverage from 0.771/0.741 to 0.845/0.947. Because each identity is supervised inside its annotated region, the loss is reduced only by placing the right person in the right location; it therefore carries spatial as well as identity information, so the layout-only gain below understates spatial grounding. P7 combines LG-ID Loss, ID tokens, Representation Forcing, Layout CoT, and text-to-layout data, reaching 0.555/0.461; these simultaneous changes prevent attribution of the margin. Its Coverage is not a threshold artifact: raising the matching threshold from 0.20 to 0.25, 0.30, and 0.35 lowers Coverage by only 0.010, 0.017, and 0.027, while just 3.6\% of references lie near the boundary (Appendix~\ref{sec:appendix-layout-score}). Lowering the threshold to 0.15 changes Coverage by only $+0.012$, further confirming robustness.

\paragraph{Effect of Layout CoT.}
P1 omits Layout CoT and ID tokens, while P2 and P3 use model-predicted and target-derived layouts, respectively. Model-predicted layout raises Sim(Ref) from 0.339 to 0.364, Count from 0.771 to 0.828, and Coverage from 0.741 to 0.813, with little Sim(Tgt) change. Ground-truth layout further raises Sim(Ref)/Sim(Tgt) to 0.412/0.367 and improves every layout metric. As a target-derived oracle rather than an achievable planning gain, it shows that \textbf{high-quality layout improves both composition and identity} and that planning quality remains a bottleneck.

\paragraph{Effect of ID tokens and Representation Forcing.}
ID tokens yield small, consistent gains, with the largest increase in Count (0.771 to 0.827); Representation Forcing further improves Sim(Ref)/Sim(Tgt) by $+0.013/+0.015$. Together with the identity-specific attention in Figure~\ref{fig:rf-attention}, these modest gains support improved identity addressability rather than a large direct identity gain.

\subsection{Analysis}
\label{sec:experiment-analysis}

\begin{figure}[t]
  \centering
  \includegraphics[width=\linewidth]{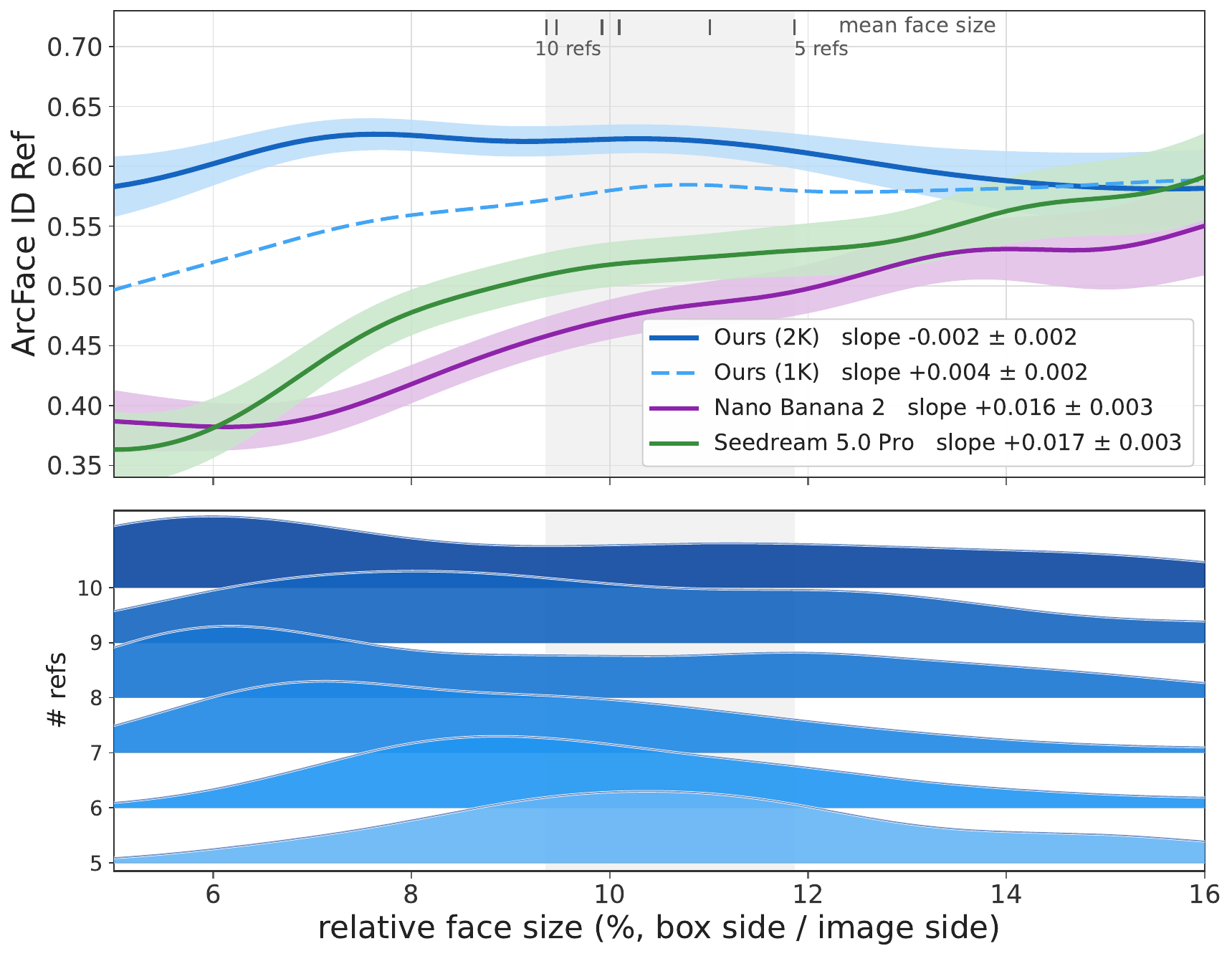}
  \caption{\small \textbf{Identity similarity against relative face size.} Face size is measured as a fraction of the image side. Curves are local means with 95\% confidence bands, annotated with the fitted slope. The grey band spans the mean face size from ten references on its left edge to five on its right. The lower panel shows the distribution of relative face size for each group size.}
  \label{fig:face-size}
  \label{fig:face-size-rel}
\end{figure}

\begin{figure}[!ht]
  \centering
  \includegraphics[width=\linewidth]{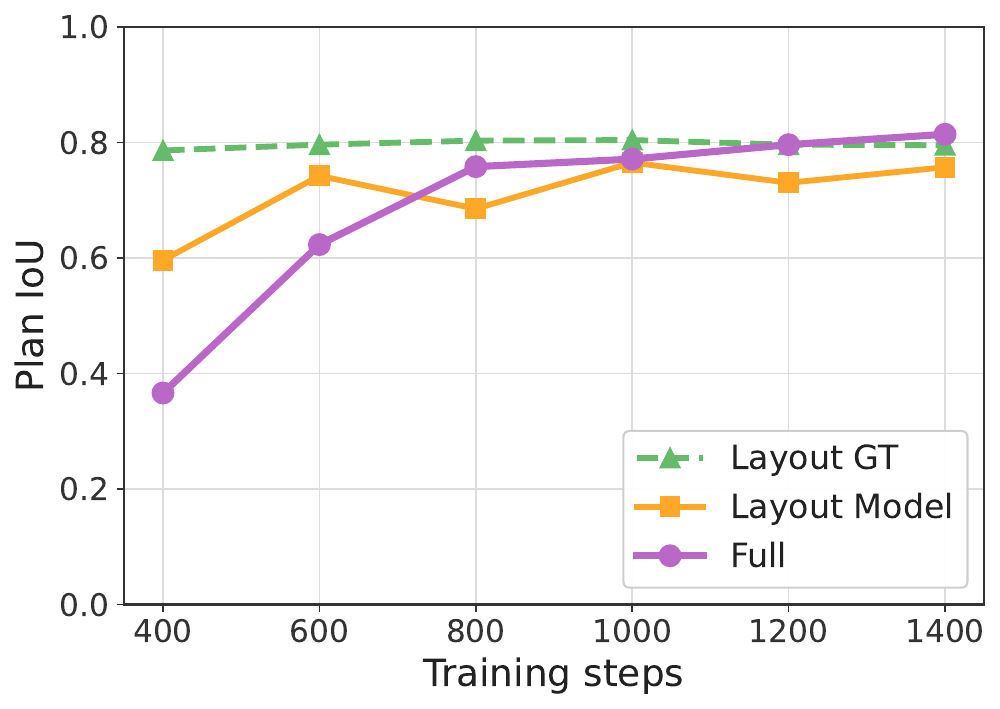}
  \caption{\small \textbf{Plan IoU convergence.} The model rapidly converges to follow the planned layout.}
  \label{fig:plan-iou}
\end{figure}

\paragraph{Planning versus execution.}
Following its own plan, the full model converges to 0.814 Plan IoU in Figure~\ref{fig:plan-iou}, matching the flat 0.79--0.80 reference of the oracle-plan variant, so it executes its own plan about as faithfully as a supplied one. Substituting that ground-truth layout nevertheless improves identity and coverage (Table~\ref{tab:ablations}), locating the residual error in plan prediction rather than execution. Since execution converges before identity preservation (Figure~\ref{fig:appendix-highres}), the main opportunity lies on the understanding side; Appendix~\ref{sec:appendix-rls} defines the Relative Layout Score (RLS), which scores plan quality against the ground-truth configuration, and Appendices~\ref{sec:appendix-high-resolution} and~\ref{sec:appendix-idloss-weight} report it.

\begin{figure}[!ht]
  \centering
  \includegraphics[width=\linewidth]{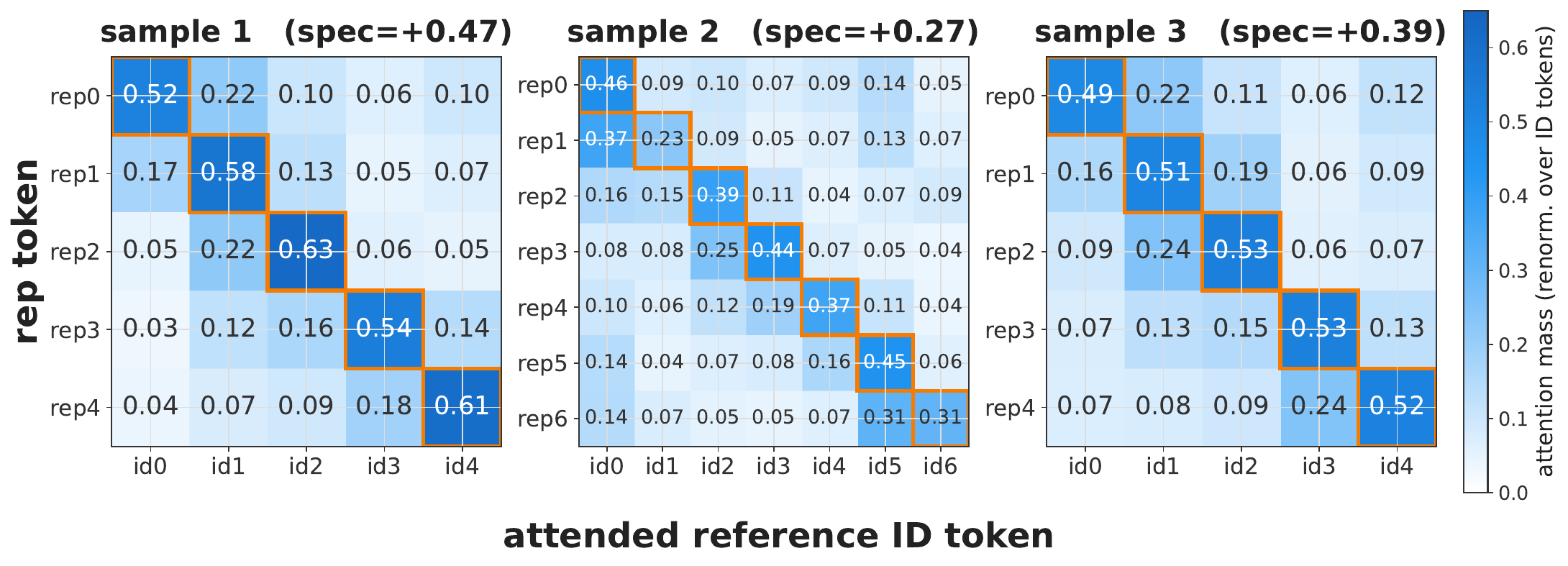}
  \caption{\small \textbf{Attention at Representation Forcing prediction positions.} In the three illustrated examples, layer-24 attention from the supervised identity predictions to reference ID tokens shows diagonal dominance.}
  \label{fig:rf-attention}
\end{figure}

\paragraph{Representation Forcing mechanism.}
Figure~\ref{fig:rf-attention} shows layer-24 diagonal attention of 0.378--0.574 versus 0.104--0.121 off-diagonal, a $+0.275$ to $+0.467$ identity-specificity gap. This supports identity-specific routing at supervised prediction positions, but not a causal effect without a no-RF control or comparison to generated face patches.

\paragraph{Face size and scaling degradation.}
Similarity could fall with group size simply because faces get smaller. Figure~\ref{fig:face-size} shows that the baselines depend strongly on face size, gaining 0.016--0.019 similarity per percentage point of relative face side, whereas our slope is nearly flat at $-0.002$. Controlling for face size removes most of the group-size slope for GPT-Image-2 and Seedream 5.0 Pro but leaves ours essentially unchanged. The reduction in face resolution therefore does not explain our remaining group-size degradation, which may reflect other factors that grow with group size, including cross-identity interference, occlusion, and compositional complexity. Within a fixed pixel band \methodname{} leads below 200~px and GPT-Image-2 overtakes it above. This pattern is qualitatively consistent with GPT-Image-2's higher Copy-Paste, although the association does not identify a causal explanation. Appendix~\ref{sec:appendix-face-size} gives the fits and the per-band numbers.

\section{Conclusion}
\label{sec:conclusion}

We presented \methodname, combining ID tokens, Representation Forcing, identity-aware Layout CoT, and LG-ID Loss for five-to-ten-person generation. On an identity-disjoint benchmark, Ours achieves the highest Sim(Tgt)/Sim(Ref) (0.538/0.641), with lower Copy-Paste than GPT-Image-2.5 (0.149 versus 0.197). Its coverage reaches 0.945, while quality and identity-collision scores leave room for improvement. LG-ID Loss provides the largest gain, Layout CoT adds coverage and spatial control, and diagnostics identify plan prediction as the remaining bottleneck within this benchmark. More broadly, multi-person identity supervision benefits from explicit spatial addresses. Appendix~\ref{sec:limitations} discusses limitations and responsible use.

\bibliographystyle{ieeenat_fullname}
\bibliography{references}

@string(CVPR= {IEEE Conf. Comput. Vis. Pattern Recog.})

@string(CVPR  = {CVPR})

@misc{ma2025hpsv3widespectrumhumanpreference,
        title={HPSv3: Towards Wide-Spectrum Human Preference Score}, 
        author={Yuhang Ma and Xiaoshi Wu and Keqiang Sun and Hongsheng Li},
        year={2025},
        eprint={2508.03789},
        archivePrefix={arXiv},
        primaryClass={cs.CV},
        url={https://arxiv.org/abs/2508.03789}, 
  }

@inproceedings{borse2026ar2can,
  title={Ar2can: An architect and an artist leveraging a canvas for multi-human generation},
  author={Borse, Shubhankar and Pham, Phuc and Farhadzadeh, Farzad and Choi, Seokeon and Nguyen, Phong and Tran, Anh and Yun, Sungrack and Hayat, Munawar and Porikli, Fatih},
  booktitle={Proceedings of the IEEE/CVF Conference on Computer Vision and Pattern Recognition},
  pages={550--560},
  year={2026}
}

@article{borse2025multihuman,
  title={MultiHuman-Testbench: Benchmarking Image Generation for Multiple Humans},
  author={Borse, Shubhankar and Choi, Seokeon and Park, Sunghyun and Kim, Jeongho and Kadambi, Shreya and Garrepalli, Risheek and Yun, Sungrack and Hayat, Munawar and Porikli, Fatih},
  journal={arXiv preprint arXiv:2506.20879},
  year={2025}
}

@inproceedings{borse2026resolving,
  title={Resolving the Identity Crisis in Text-to-Image Generation},
  author={Borse, Shubhankar and Farhadzadeh, Farzad and Hayat, Munawar and Porikli, Fatih},
  booktitle={Proceedings of the IEEE/CVF Conference on Computer Vision and Pattern Recognition},
  pages={36703--36712},
  year={2026}
}

@article{xu2025withanyone,
  title={WithAnyone: Towards Controllable and ID-Consistent Image Generation}, 
  author={Hengyuan Xu and Wei Cheng and Peng Xing and Yixiao Fang and Shuhan Wu and Rui Wang and Xianfang Zeng and Gang Yu and Xinjun Ma and Yu-Gang Jiang},
  journal={arXiv preprint arxiv:2510.14975},
  year={2025}
}

@inproceedings{valevski2023face0,
  title={Face0: Instantaneously conditioning a text-to-image model on a face},
  author={Valevski, Dani and Lumen, Danny and Matias, Yossi and Leviathan, Yaniv},
  booktitle={SIGGRAPH Asia 2023 Conference Papers},
  pages={1--10},
  year={2023}
}

@article{wang2024stableidentity,
  title={Stableidentity: Inserting anybody into anywhere at first sight},
  author={Wang, Qinghe and Jia, Xu and Li, Xiaomin and Li, Taiqing and Ma, Liqian and Zhuge, Yunzhi and Lu, Huchuan},
  journal={arXiv preprint arXiv:2401.15975},
  year={2024}
}

@article{yan2023facestudio,
  title={Facestudio: Put your face everywhere in seconds},
  author={Yan, Yuxuan and Zhang, Chi and Wang, Rui and Zhou, Yichao and Zhang, Gege and Cheng, Pei and Yu, Gang and Fu, Bin},
  journal={arXiv preprint arXiv:2312.02663},
  year={2023}
}

@article{xiao2025fastcomposer,
  title={Fastcomposer: Tuning-free multi-subject image generation with localized attention},
  author={Xiao, Guangxuan and Yin, Tianwei and Freeman, William T and Durand, Fr{\'e}do and Han, Song},
  journal={International Journal of Computer Vision},
  volume={133},
  number={3},
  pages={1175--1194},
  year={2025},
  publisher={Springer}
}

@article{he2024uniportrait,
  title={UniPortrait: A Unified Framework for Identity-Preserving Single-and Multi-Human Image Personalization},
  author={He, Junjie and Geng, Yifeng and Bo, Liefeng},
  journal={arXiv preprint arXiv:2408.05939},
  year={2024}
}

@article{wu2025uno,
  title={Less-to-more generalization: Unlocking more controllability by in-context generation},
  author={Wu, Shaojin and Huang, Mengqi and Wu, Wenxu and Cheng, Yufeng and Ding, Fei and He, Qian},
  journal={arXiv preprint arXiv:2504.02160},
  year={2025}
}

@InProceedings{zhang2025idpatch,
    author    = {Zhang, Yimeng and Zhi, Tiancheng and Liu, Jing and Sang, Shen and Jiang, Liming and Yan, Qing and Liu, Sijia and Luo, Linjie},
    title     = {ID-Patch: Robust ID Association for Group Photo Personalization},
    booktitle = {Proceedings of the IEEE/CVF Conference on Computer Vision and Pattern Recognition (CVPR)},
    month     = {June},
    year      = {2025}
}

@InProceedings{guo2024pulid,
  title={PuLID: Pure and Lightning ID Customization via Contrastive Alignment},
  author={Guo, Zinan and Wu, Yanze and Chen, Zhuowei and Chen, Lang and Zhang, Peng and He, Qian},
  booktitle={Advances in Neural Information Processing Systems},
  year={2024}
}

@article{chen2025xverse,
  title={XVerse: Consistent Multi-Subject Control of Identity and Semantic Attributes via DiT Modulation},
  author={Chen, Bowen and Zhao, Mengyi and Sun, Haomiao and Chen, Li and Wang, Xu and Du, Kang and Wu, Xinglong},
  journal={arXiv preprint arXiv:2506.21416},
  year={2025}


}

@inproceedings{li2024photomaker,
  title={Photomaker: Customizing realistic human photos via stacked id embedding},
  author={Li, Zhen and Cao, Mingdeng and Wang, Xintao and Qi, Zhongang and Cheng, Ming-Ming and Shan, Ying},
  booktitle={Proceedings of the IEEE/CVF conference on computer vision and pattern recognition},
  pages={8640--8650},
  year={2024}
}

@article{ye2023ipadapter,
  title={IP-Adapter: Text Compatible Image Prompt Adapter for Text-to-Image Diffusion Models},
  author={Ye, Hu and Zhang, Jun and Liu, Sibo and Han, Xiao and Yang, Wei},
  booktitle={arXiv preprint arxiv:2308.06721},
  year={2023}
}

@article{wang2024instantid,
  title={InstantID: Zero-shot Identity-Preserving Generation in Seconds},
  author={Wang, Qixun and Bai, Xu and Wang, Haofan and Qin, Zekui and Chen, Anthony},
  journal={arXiv preprint arXiv:2401.07519},
  year={2024}
}

@inproceedings{papantoniou2024arc2face,
  title={Arc2face: A foundation model for id-consistent human faces},
  author={Papantoniou, Foivos Paraperas and Lattas, Alexandros and Moschoglou, Stylianos and Deng, Jiankang and Kainz, Bernhard and Zafeiriou, Stefanos},
  booktitle={European Conference on Computer Vision},
  pages={241--261},
  year={2024},
  organization={Springer}
}

@article{kim2024instantfamily,
  title={Instantfamily: Masked attention for zero-shot multi-id image generation},
  author={Kim, Chanran and Lee, Jeongin and Joung, Shichang and Kim, Bongmo and Baek, Yeul-Min},
  journal={arXiv preprint arXiv:2404.19427},
  year={2024}
}

@article{jiang2025infiniteyou,
  title={InfiniteYou: Flexible photo recrafting while preserving your identity},
  author={Jiang, Liming and Yan, Qing and Jia, Yumin and Liu, Zichuan and Kang, Hao and Lu, Xin},
  journal={arXiv preprint arXiv:2503.16418},
  year={2025}
}

@article{mou2025dreamo,
  title={DreamO: A Unified Framework for Image Customization},
  author={Mou, Chong and Wu, Yanze and Wu, Wenxu and Guo, Zinan and Zhang, Pengze and Cheng, Yufeng and Luo, Yiming and Ding, Fei and Zhang, Shiwen and Li, Xinghui and others},
  journal={arXiv preprint arXiv:2504.16915},
  year={2025}
}

@article{guo2025musar,
  title={MUSAR: Exploring Multi-Subject Customization from Single-Subject Dataset via Attention Routing},
  author={Guo, Zinan and Zhang, Pengze and Wu, Yanze and Mou, Chong and Zhao, Songtao and He, Qian},
  journal={arXiv preprint arXiv:2505.02823},
  year={2025}
}

@article{wu2025omnigen2,
  title={OmniGen2: Exploration to Advanced Multimodal Generation},
  author={Chenyuan Wu and Pengfei Zheng and Ruiran Yan and Shitao Xiao and Xin Luo and Yueze Wang and Wanli Li and Xiyan Jiang and Yexin Liu and Junjie Zhou and Ze Liu and Ziyi Xia and Chaofan Li and Haoge Deng and Jiahao Wang and Kun Luo and Bo Zhang and Defu Lian and Xinlong Wang and Zhongyuan Wang and Tiejun Huang and Zheng Liu},
  journal={arXiv preprint arXiv:2506.18871},
  year={2025}
}

@article{wu2025uso,
    title={USO: Unified Style and Subject-Driven Generation via Disentangled and Reward Learning},
    author={Shaojin Wu and Mengqi Huang and Yufeng Cheng and Wenxu Wu and Jiahe Tian and Yiming Luo and Fei Ding and Qian He},
    year={2025},
    eprint={2508.18966},
    archivePrefix={arXiv},
    primaryClass={cs.CV},
}

@article{cheng2025umo,
  title={UMO: Scaling Multi-Identity Consistency for Image Customization via Matching Reward},
  author={Cheng, Yufeng and Wu, Wenxu and Wu, Shaojin and Huang, Mengqi and Ding, Fei and He, Qian},
  journal={arXiv preprint arXiv:2509.06818},
  year={2025}
}

@inproceedings{peng2024portraitbooth,
  title={Portraitbooth: A versatile portrait model for fast identity-preserved personalization},
  author={Peng, Xu and Zhu, Junwei and Jiang, Boyuan and Tai, Ying and Luo, Donghao and Zhang, Jiangning and Lin, Wei and Jin, Taisong and Wang, Chengjie and Ji, Rongrong},
  booktitle={Proceedings of the IEEE/CVF Conference on Computer Vision and Pattern Recognition},
  pages={27080--27090},
  year={2024}
}

@article{jiang2026t2i,
  title={T2i-r1: Reinforcing image generation with collaborative semantic-level and token-level cot},
  author={Jiang, Dongzhi and Guo, Ziyu and Zhang, Renrui and Zong, Zhuofan and Li, Hao and Zhuo, Le and Yan, Shilin and Heng, Pheng-Ann and Li, Hongsheng},
  journal={Advances in Neural Information Processing Systems},
  volume={38},
  pages={39856--39890},
  year={2026}
}

@article{lian2023llm,
  title={Llm-grounded diffusion: Enhancing prompt understanding of text-to-image diffusion models with large language models},
  author={Lian, Long and Li, Boyi and Yala, Adam and Darrell, Trevor},
  journal={arXiv preprint arXiv:2305.13655},
  year={2023}
}

@article{feng2023layoutgpt,
  title={Layoutgpt: Compositional visual planning and generation with large language models},
  author={Feng, Weixi and Zhu, Wanrong and Fu, Tsu-jui and Jampani, Varun and Akula, Arjun and He, Xuehai and Basu, Sugato and Wang, Xin Eric and Wang, William Yang},
  journal={Advances in Neural Information Processing Systems},
  volume={36},
  pages={18225--18250},
  year={2023}
}

@inproceedings{he2025plangen,
  title={Plangen: Towards unified layout planning and image generation in auto-regressive vision language models},
  author={He, Runze and Cheng, Bo and Ma, Yuhang and Jia, Qingxiang and Liu, Shanyuan and Ma, Ao and Wu, Xiaoyu and Wu, Liebucha and Leng, Dawei and Yin, Yuhui},
  booktitle={Proceedings of the IEEE/CVF International Conference on Computer Vision},
  pages={18143--18154},
  year={2025}
}

@inproceedings{gupta2023visual,
  title={Visual programming: Compositional visual reasoning without training},
  author={Gupta, Tanmay and Kembhavi, Aniruddha},
  booktitle={Proceedings of the IEEE/CVF conference on computer vision and pattern recognition},
  pages={14953--14962},
  year={2023}
}

@inproceedings{yang2024mastering,
  title={Mastering Text-to-Image Diffusion: Recaptioning, Planning, and Generating with Multimodal LLMs.},
  author={Yang, Ling and Yu, Zhaochen and Meng, Chenlin and Xu, Minkai and Ermon, Stefano and Cui, Bin},
  booktitle={Icml},
  volume={3},
  number={6},
  pages={7},
  year={2024}
}

@article{zhang2024realcompo,
  title={Realcompo: Balancing realism and compositionality improves text-to-image diffusion models},
  author={Zhang, Xinchen and Yang, Ling and Cai, Yaqi and Yu, Zhaochen and Wang, Kai-Ni and Xie, Jiake and Tian, Ye and Xu, Minkai and Tang, Yong and Yang, Yujiu and others},
  journal={Advances in Neural Information Processing Systems},
  volume={37},
  pages={96963--96992},
  year={2024}
}

@article{liu2026atlas,
  title={Think, Plan, Paint: Layout-Aware Reasoning for Controllable Image Generation in Unified Models},
  author={Liu, Junhao and Zhang, Jian-Wei and Huang, Tao and Yang, Miles and Zhong, Zhao and Bo, Liefeng},
  journal={arXiv preprint arXiv:2607.16409},
  year={2026}
}

@article{fang2025got,
  title={Got: Unleashing reasoning capability of multimodal large language model for visual generation and editing},
  author={Fang, Rongyao and Duan, Chengqi and Wang, Kun and Huang, Linjiang and Li, Hao and Yan, Shilin and Tian, Hao and Zeng, Xingyu and Zhao, Rui and Dai, Jifeng and others},
  journal={arXiv preprint arXiv:2503.10639},
  year={2025}
}

@article{duan2025got,
  title={Got-r1: Unleashing reasoning capability of mllm for visual generation with reinforcement learning},
  author={Duan, Chengqi and Fang, Rongyao and Wang, Yuqing and Wang, Kun and Huang, Linjiang and Zeng, Xingyu and Li, Hongsheng and Liu, Xihui},
  journal={arXiv preprint arXiv:2505.17022},
  year={2025}
}

@article{wang2026representation,
  title={Representation Forcing for Bottleneck-Free Unified Multimodal Models},
  author={Wang, Yuqing and Lin, Zhijie and Yang, Ceyuan and Zhao, Yang and Xiao, Fei and He, Hao and Zhao, Qi and Ding, Zihan and Wang, Fuyun and Wang, Shuai and others},
  journal={arXiv preprint arXiv:2605.31604},
  year={2026}
}

@article{liu2025muon,
  title={Muon is scalable for llm training},
  author={Liu, Jingyuan and Su, Jianlin and Yao, Xingcheng and Jiang, Zhejun and Lai, Guokun and Du, Yulun and Qin, Yidao and Xu, Weixin and Lu, Enzhe and Yan, Junjie and others},
  journal={arXiv preprint arXiv:2502.16982},
  year={2025}
}

@inproceedings{radford2021learning,
  title={Learning transferable visual models from natural language supervision},
  author={Radford, Alec and Kim, Jong Wook and Hallacy, Chris and Ramesh, Aditya and Goh, Gabriel and Agarwal, Sandhini and Sastry, Girish and Askell, Amanda and Mishkin, Pamela and Clark, Jack and others},
  booktitle={International conference on machine learning},
  pages={8748--8763},
  year={2021},
  organization={PmLR}
}

@misc{flux-2-2025,
    author={Black Forest Labs},
    title={{FLUX.2: Frontier Visual Intelligence}},
    year={2025},
    howpublished={\url{https://bfl.ai/blog/flux-2}},
}

@article{diao2026sensenova,
  title={Sensenova-u1: Unifying multimodal understanding and generation with neo-unify architecture},
  author={Diao, Haiwen and Wu, Penghao and Deng, Hanming and Wang, Jiahao and Bai, Shihao and Wu, Silei and Fan, Weichen and Ye, Wenjie and Tong, Wenwen and Fan, Xiangyu and others},
  journal={arXiv preprint arXiv:2605.12500},
  year={2026}
}

@article{cai2026hidream,
  title={Hidream-o1-image: A natively unified image generative foundation model with pixel-level unified transformer},
  author={Cai, Qi and Chen, Jingwen and Gao, Chengmin and Gong, Zijian and Li, Yehao and Pan, Yingwei and Peng, Yi and Qiu, Zhaofan and Yu, Kai and Zhang, Yiheng and others},
  journal={arXiv preprint arXiv:2605.11061},
  year={2026}
}

@article{team2025longcat,
  title={Longcat-image technical report},
  author={Team, Meituan LongCat and Ma, Hanghang and Tan, Haoxian and Huang, Jiale and Wu, Junqiang and He, Jun-Yan and Gao, Lishuai and Xiao, Songlin and Wei, Xiaoming and Ma, Xiaoqi and others},
  journal={arXiv preprint arXiv:2512.07584},
  year={2025}
}

@misc{yolo11_ultralytics,
  author  = {Ultralytics},
  title   = {Ultralytics YOLO11},
  version = {11.0.0},
  year    = {2024},
  url     = {https://github.com/ultralytics/ultralytics},
  orcid   = {0000-0001-5950-6979, 0000-0003-3783-7069},
  license = {AGPL-3.0}
}

@article{deng2025bagel,
  title   = {Emerging Properties in Unified Multimodal Pretraining},
  author  = {Deng, Chaorui and Zhu, Deyao and Li, Kunchang and Gou, Chenhui and Li, Feng and Wang, Zeyu and Zhong, Shu and Yu, Weihao and Nie, Xiaonan and Song, Ziang and Shi, Guang and Fan, Haoqi},
  journal = {arXiv preprint arXiv:2505.14683},
  year    = {2025}
}

@article{cao2025hunyuanimage,
  title={Hunyuanimage 3.0 technical report},
  author={Cao, Siyu and Chen, Hangting and Chen, Peng and Cheng, Yiji and Cui, Yutao and Deng, Xinchi and Dong, Ying and Gong, Kipper and Gu, Tianpeng and Gu, Xiusen and others},
  journal={arXiv preprint arXiv:2509.23951},
  year={2025}
}

@inproceedings{zhou2025transfusion,
  title={Transfusion: Predict the next token and diffuse images with one multi-modal model},
  author={Zhou, Chunting and Yu, Lili and Babu, Arun and Tirumala, Kushal and Yasunaga, Michihiro and Shamis, Leonid and Kahn, Jacob and Ma, Xuezhe and Zettlemoyer, Luke and Levy, Omer},
  booktitle={International Conference on Learning Representations},
  volume={2025},
  pages={6446--6469},
  year={2025}
}

@article{lipman2022flow,
  title={Flow matching for generative modeling},
  author={Lipman, Yaron and Chen, Ricky TQ and Ben-Hamu, Heli and Nickel, Maximilian and Le, Matt},
  journal={arXiv preprint arXiv:2210.02747},
  year={2022}
}

@misc{openai2026gptimage2,
  author    = {OpenAI},
  title     = {Introducing ChatGPT Images 2.0},
  year      = {2026},
  month     = apr,
  publisher = {OpenAI},
  url       = {https://openai.com/index/introducing-chatgpt-images-2-0/}
}

@misc{raisinghani2025nanobananapro,
  author    = {{Google DeepMind}},
  title     = {Introducing Nano Banana Pro},
  year      = {2025},
  month     = nov,
  publisher = {Google DeepMind},
  url       = {https://blog.google/innovation-and-ai/products/nano-banana-pro/}
}

@misc{raisinghani2026nanobanana2,
  author    = {{Google DeepMind}},
  title     = {Nano Banana 2: Combining Pro Capabilities with Lightning-Fast Speed},
  year      = {2026},
  month     = feb,
  publisher = {Google DeepMind},
  url       = {https://blog.google/innovation-and-ai/technology/ai/nano-banana-2/}
}

@misc{bytedanceseed2026seedream5pro,
  author    = {{ByteDance Seed Team}},
  title     = {Beyond Generation, It Understands Design: Introducing Seedream 5.0 Pro},
  year      = {2026},
  month     = jul,
  publisher = {ByteDance},
  url       = {https://seed.bytedance.com/en/blog/beyond-generation-it-understands-design-introducing-seedream-5-0-pro}
}

@misc{openai2025gpt4o,
  title={Addendum to GPT-4o System Card: Native image generation},
  author={OpenAI},
  year={2025},
  publisher={OpenAI},
  url={https://cdn.openai.com/11998be9-5319-4302-bfbf-1167e093f1fb/Native_Image_Generation_System_Card.pdf}
}

@misc{flux2024,
    author={Black Forest Labs},
    title={FLUX},
    year={2024},
    howpublished={\url{https://github.com/black-forest-labs/flux}},
}

@misc{fluxkrea,
    author={Black Forest Labs},
    title={FLUX.1 Krea},
    year={2025},
    howpublished={\url{https://huggingface.co/black-forest-labs/FLUX.1-Krea-dev}},
}

@misc{wu2025qwenimage,
      title={Qwen-Image Technical Report}, 
      author={Chenfei Wu and Jiahao Li and Jingren Zhou and Junyang Lin and Kaiyuan Gao and Kun Yan and Sheng-ming Yin and Shuai Bai and Xiao Xu and Yilei Chen and Yuxiang Chen and Zecheng Tang and Zekai Zhang and Zhengyi Wang and An Yang and Bowen Yu and Chen Cheng and Dayiheng Liu and Deqing Li and Hang Zhang and Hao Meng and Hu Wei and Jingyuan Ni and Kai Chen and Kuan Cao and Liang Peng and Lin Qu and Minggang Wu and Peng Wang and Shuting Yu and Tingkun Wen and Wensen Feng and Xiaoxiao Xu and Yi Wang and Yichang Zhang and Yongqiang Zhu and Yujia Wu and Yuxuan Cai and Zenan Liu},
      year={2025},
      eprint={2508.02324},
      archivePrefix={arXiv},
      primaryClass={cs.CV},
      url={https://arxiv.org/abs/2508.02324}, 
}

@article{batifol2025flux,
  title={FLUX. 1 Kontext: Flow Matching for In-Context Image Generation and Editing in Latent Space},
  author={Batifol, Stephen and Blattmann, Andreas and Boesel, Frederic and Consul, Saksham and Diagne, Cyril and Dockhorn, Tim and English, Jack and English, Zion and Esser, Patrick and Kulal, Sumith and others},
  journal={arXiv e-prints},
  pages={arXiv--2506},
  year={2025}
}

@inproceedings{peebles2023scalable,
  title={Scalable diffusion models with transformers},
  author={Peebles, William and Xie, Saining},
  booktitle={Proceedings of the IEEE/CVF international conference on computer vision},
  pages={4195--4205},
  year={2023}
}

@article{oord2018representation,
  title={Representation learning with contrastive predictive coding},
  author={Oord, Aaron van den and Li, Yazhe and Vinyals, Oriol},
  journal={arXiv preprint arXiv:1807.03748},
  year={2018}
}

@article{xiao2024omnigen,
  title={Omnigen: Unified image generation},
  author={Xiao, Shitao and Wang, Yueze and Zhou, Junjie and Yuan, Huaying and Xing, Xingrun and Yan, Ruiran and Li, Chaofan and Wang, Shuting and Huang, Tiejun and Liu, Zheng},
  journal={arXiv preprint arXiv:2409.11340},
  year={2024}
}

@article{liu2025step1x,
  title={Step1X-Edit: A Practical Framework for General Image Editing},
  author={Liu, Shiyu and Han, Yucheng and Xing, Peng and Yin, Fukun and Wang, Rui and Cheng, Wei and Liao, Jiaqi and Wang, Yingming and Fu, Honghao and Han, Chunrui and others},
  journal={arXiv preprint arXiv:2504.17761},
  year={2025}
}

@inproceedings{stablediffusion,
  title={High-resolution image synthesis with latent diffusion models},
  author={Rombach, Robin and Blattmann, Andreas and Lorenz, Dominik and Esser, Patrick and Ommer, Bj{\"o}rn},
  booktitle={Proceedings of the IEEE/CVF conference on computer vision and pattern recognition},
  pages={10684--10695},
  year={2022}
}

@misc{aesthetic-predictor-v2-5,
  author       = {discus0434},
  title        = {aesthetic-predictor-v2-5},
  year         = {2023},
  howpublished = {\url{https://github.com/discus0434/aesthetic-predictor-v2-5}},
  note         = {Accessed: 2025-05-12}
}

@inproceedings{Deng2020retinaface,
title = {RetinaFace: Single-Shot Multi-Level Face Localisation in the Wild},
author = {Deng, Jiankang and Guo, Jia and Ververas, Evangelos and Kotsia, Irene and Zafeiriou, Stefanos},
booktitle = {CVPR},
year = {2020}
}

@inproceedings{zhai2023siglip,
  title={Sigmoid loss for language image pre-training},
  author={Zhai, Xiaohua and Mustafa, Basil and Kolesnikov, Alexander and Beyer, Lucas},
  booktitle={Proceedings of the IEEE/CVF international conference on computer vision},
  pages={11975--11986},
  year={2023}
}

@inproceedings{schroff2015facenet,
  title={Facenet: A unified embedding for face recognition and clustering},
  author={Schroff, Florian and Kalenichenko, Dmitry and Philbin, James},
  booktitle={Proceedings of the IEEE conference on computer vision and pattern recognition},
  pages={815--823},
  year={2015}
}

@misc{oquab2023dinov2,
  title={DINOv2: Learning Robust Visual Features without Supervision},
  author={Oquab, Maxime and Darcet, Timothée and Moutakanni, Theo and Vo, Huy V. and Szafraniec, Marc and Khalidov, Vasil and Fernandez, Pierre and Haziza, Daniel and Massa, Francisco and El-Nouby, Alaaeldin and Howes, Russell and Huang, Po-Yao and Xu, Hu and Sharma, Vasu and Li, Shang-Wen and Galuba, Wojciech and Rabbat, Mike and Assran, Mido and Ballas, Nicolas and Synnaeve, Gabriel and Misra, Ishan and Jegou, Herve and Mairal, Julien and Labatut, Patrick and Joulin, Armand and Bojanowski, Piotr},
  journal={arXiv:2304.07193},
  year={2023}
}

@inproceedings{kim2022adaface,
  title={Adaface: Quality adaptive margin for face recognition},
  author={Kim, Minchul and Jain, Anil K and Liu, Xiaoming},
  booktitle={Proceedings of the IEEE/CVF conference on computer vision and pattern recognition},
  pages={18750--18759},
  year={2022}
}

@inproceedings{deng2019arcface,
    title={Arcface: Additive angular margin loss for deep face recognition},
    author={Deng, Jiankang and Guo, Jia and Xue, Niannan and Zafeiriou, Stefanos},
    booktitle={Proceedings of the IEEE/CVF Conference on Computer Vision and Pattern Recognition},
    pages={4690--4699},
    year={2019}
    }

@article{ho2020ddpm,
  title={Denoising diffusion probabilistic models},
  author={Ho, Jonathan and Jain, Ajay and Abbeel, Pieter},
  journal={Advances in neural information processing systems},
  volume={33},
  pages={6840--6851},
  year={2020}
}

@inproceedings{ronneberger2015u,
  title={U-net: Convolutional networks for biomedical image segmentation},
  author={Ronneberger, Olaf and Fischer, Philipp and Brox, Thomas},
  booktitle={Medical image computing and computer-assisted intervention--MICCAI 2015: 18th international conference, Munich, Germany, October 5-9, 2015, proceedings, part III 18},
  pages={234--241},
  year={2015},
  organization={Springer}
}

@article{vaswani2017attention,
  title={Attention is all you need},
  author={Vaswani, Ashish and Shazeer, Noam and Parmar, Niki and Uszkoreit, Jakob and Jones, Llion and Gomez, Aidan N and Kaiser, {\L}ukasz and Polosukhin, Illia},
  journal={Advances in neural information processing systems},
  volume={30},
  year={2017}
}

@article{dosovitskiy2020image,
  title={An image is worth 16x16 words: Transformers for image recognition at scale},
  author={Dosovitskiy, Alexey and Beyer, Lucas and Kolesnikov, Alexander and Weissenborn, Dirk and Zhai, Xiaohua and Unterthiner, Thomas and Dehghani, Mostafa and Minderer, Matthias and Heigold, Georg and Gelly, Sylvain and others},
  journal={arXiv preprint arXiv:2010.11929},
  year={2020}
}

@inproceedings{zhang2023adding,
  title={Adding conditional control to text-to-image diffusion models},
  author={Zhang, Lvmin and Rao, Anyi and Agrawala, Maneesh},
  booktitle={Proceedings of the IEEE/CVF international conference on computer vision},
  pages={3836--3847},
  year={2023}
}

@inproceedings{wang2024moa,
  title={Moa: Mixture-of-attention for subject-context disentanglement in personalized image generation},
  author={Wang, Kuan-Chieh and Ostashev, Daniil and Fang, Yuwei and Tulyakov, Sergey and Aberman, Kfir},
  booktitle={SIGGRAPH Asia 2024 Conference Papers},
  pages={1--12},
  year={2024}
}

@inproceedings{xu2024permutation,
  title={Permutation equivariance of transformers and its applications},
  author={Xu, Hengyuan and Xiang, Liyao and Ye, Hangyu and Yao, Dixi and Chu, Pengzhi and Li, Baochun},
  booktitle={Proceedings of the IEEE/CVF Conference on Computer Vision and Pattern Recognition},
  pages={5987--5996},
  year={2024}
}

@misc{openai2026gptimage25,
  author    = {{OpenAI}},
  title     = {Introducing {ChatGPT Images 2.5}},
  year      = {2026},
  publisher = {OpenAI},
  url       = {https://openai.com/index/introducing-chatgpt-images-2-5/}
}

\clearpage
\appendix
\section{Additional Qualitative Results}
\label{app:gallery}
\begin{figure*}[p]
  \centering
  \includegraphics[width=0.95\linewidth]{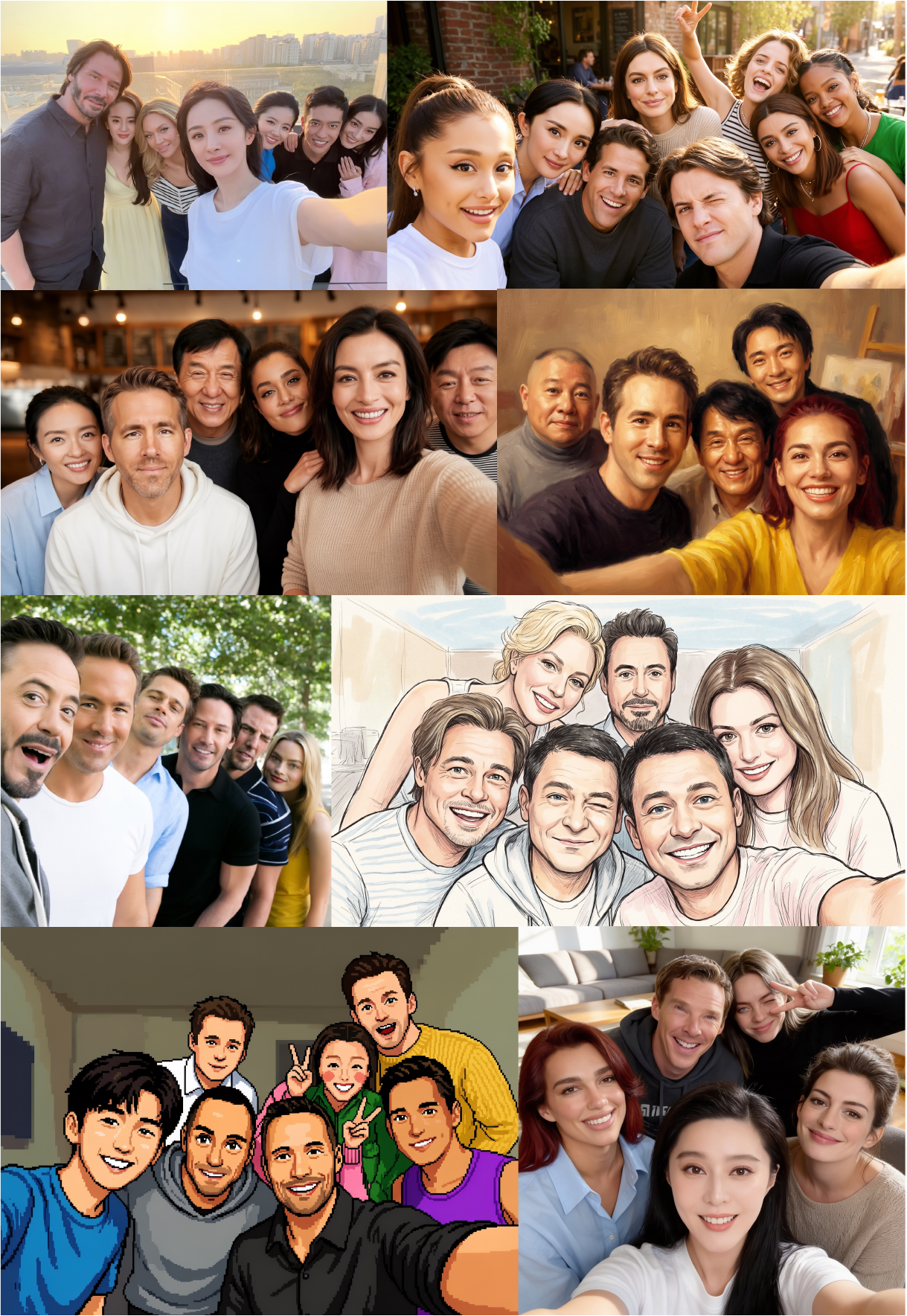}
  \caption{\small \textbf{Additional group images generated by \methodname.}}
  \label{fig:gallery3}
\end{figure*}

Figures~\ref{fig:gallery1}--\ref{fig:gallery3} collect a small selection of group images generated by \methodname. A larger anonymous gallery, in which each generated group can be inspected together with its reference identities, is included in the supplementary ZIP. After extracting the archive, open \texttt{showcase/index.html} in a browser; all gallery assets are included locally.

\ifdefined\gallerysinbody\else
\begin{figure*}[p]
  \centering
  \includegraphics[width=0.95\linewidth]{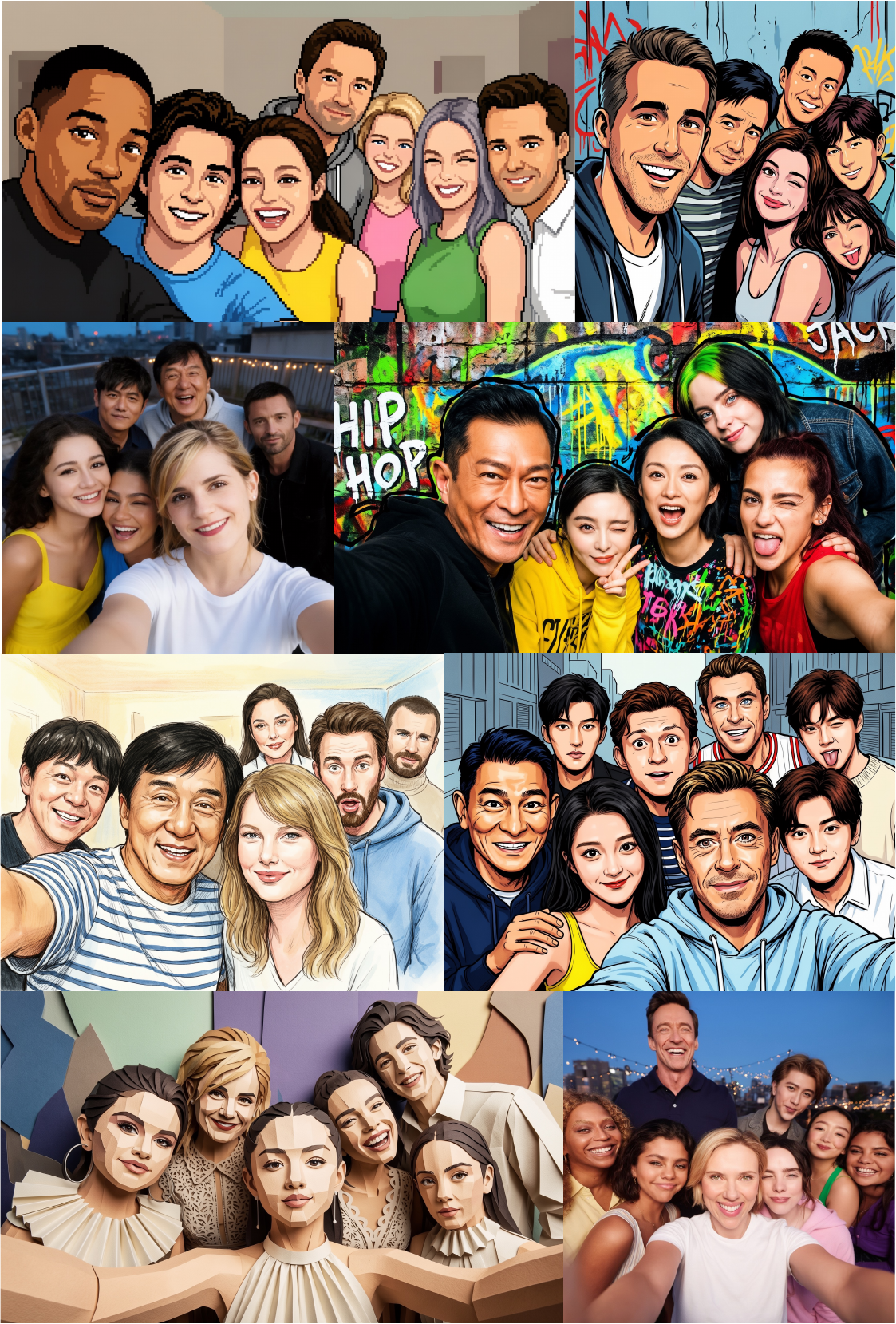}
  \caption{\small \textbf{Group images generated by \methodname.} The samples cover different group sizes, scene types, and visual styles.}
  \label{fig:gallery1}
\end{figure*}

\fi
\begin{figure*}[p]
  \centering
  \includegraphics[width=0.95\linewidth]{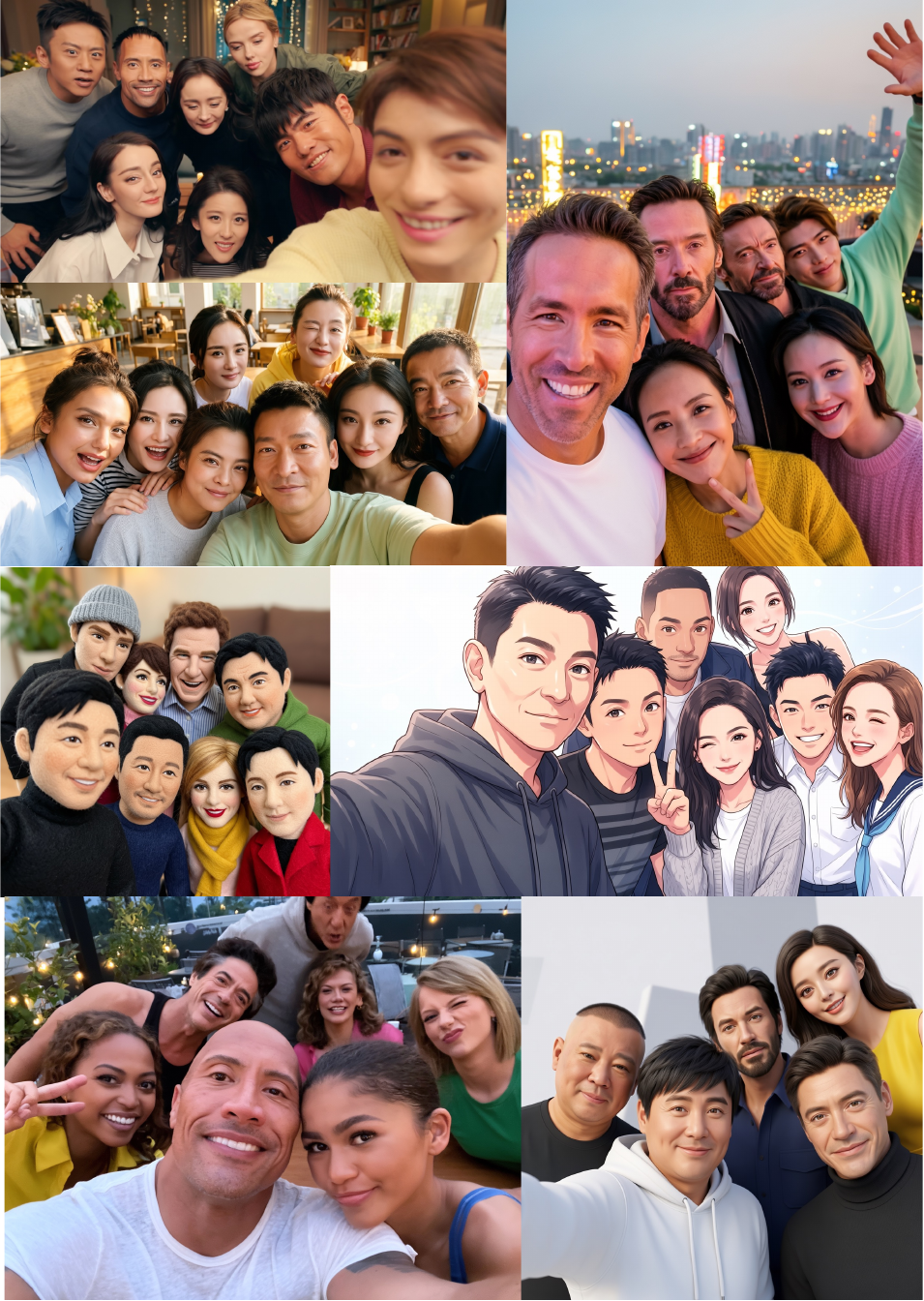}
  \caption{\small \textbf{Additional group images generated by \methodname.}}
  \label{fig:gallery2}
\end{figure*}

\section{Additional Experiments and Analysis}
\label{sec:appendix}

\subsection{Statistical Reliability of the Earlier Checkpoint}
\label{sec:appendix-reliability}

This section preserves the earlier-checkpoint analysis. Its means, intervals, and tests do not describe Ours or the updated Table~\ref{tab:main-results}; updated paired tests remain pending.

\begin{table*}[t]
  \centering
  \caption{\small \textbf{Dispersion of the earlier comparison.} Per-example means on the 210 benchmark examples with their standard errors. The last column is the single-encoder ArcFace similarity to the references for the earlier checkpoint.}
  \label{tab:appendix-reliability}
  \small
  \begin{tabular}{lcccc}
    \toprule
    Method & Sim(Tgt) $\uparrow$ & Sim(Ref) $\uparrow$ & CLIP-I $\uparrow$ & ArcFace Sim(Ref) $\uparrow$ \\
    \midrule
    \methodname{} & $0.499 \pm 0.004$ & $0.540 \pm 0.004$ & $0.861 \pm 0.004$ & $0.614 \pm 0.005$ \\
    GPT-Image-2 & $0.462 \pm 0.004$ & $0.583 \pm 0.008$ & $0.853 \pm 0.004$ & $0.566 \pm 0.009$ \\
    Nano Banana 2 & $0.451 \pm 0.005$ & $0.480 \pm 0.007$ & $0.860 \pm 0.004$ & $0.471 \pm 0.008$ \\
    Seedream 5.0 Pro & $0.436 \pm 0.005$ & $0.522 \pm 0.008$ & $0.850 \pm 0.004$ & $0.506 \pm 0.009$ \\
    \bottomrule
  \end{tabular}
\end{table*}

The earlier comparison was reported over the 210-example benchmark. To show how precisely those means are determined, Table~\ref{tab:appendix-reliability} recomputes them per example and gives the standard error for \methodname{} and the three strongest baselines. The historical paired statistics below are retained for the earlier checkpoint and require a cohort audit before reuse; they provide no significance evidence for the updated main comparison. Our Sim(Tgt) advantage over GPT-Image-2 is $+0.038$ with a 20{,}000-sample bootstrap interval of $[+0.027,+0.048]$, a Wilcoxon $p$ of $3.8\times10^{-11}$, and a higher score on 73\% of examples. Our CLIP-I margin over Nano Banana 2 is $+0.001$ with an interval of $[-0.006,+0.009]$, a $p$ of 0.98, and a higher score on 48\% of examples: the two systems are statistically indistinguishable on this metric, and this historical near-tie should not be read as a lead. Against Seedream 5.0 Pro, Sim(Ref) improves by $+0.018$ $([+0.005,+0.039])$ while Copy-Paste falls by $0.059$ $([-0.085,-0.041])$ and ArcFace similarity rises by $+0.108$ $([+0.092,+0.131])$, so both directions hold simultaneously.

\subsection{Extended Reading of the Main Comparison}
\label{sec:appendix-main-reading}

Table~\ref{tab:main-results} now evaluates Ours. It leads Sim(Tgt)/Sim(Ref) at 0.538/0.641, followed by GPT-Image-2.5 at 0.472/0.615; Copy-Paste is 0.149 versus 0.197. Excluding the ground-truth-layout methods WithAnyone and ID-Patch, Ours ranks first in Coverage (0.945) and third in Dup (0.096); it does not have the lowest collision rate. CLIP-I/DINO-I are 0.849/0.724, below Nano Banana 2 at 0.860/0.731. Aes/HPSv3 are 5.21/11.915, below GPT-Image-2.5 at 5.63/12.023. The comparisons describe available-score means; no updated significance claim is made. The group-size and face-size analyses below retain the earlier checkpoint.

In the group-size breakdown of Figure~\ref{fig:scaling}, the per-example distribution of \methodname{} is also more concentrated, whereas GPT-Image-2 shows substantially heavier upper and lower tails. Because the groups are small, especially at the upper end with 20 and 10 examples for nine and ten references, we read the figure as a consistent advantage across the whole five-to-ten range and a flatter degradation trend, rather than as a claim about any individual group size.

\subsection{Face Size and Scaling Degradation}
\label{sec:appendix-face-size}

\begin{figure}[t]
  \centering
  \includegraphics[width=\linewidth]{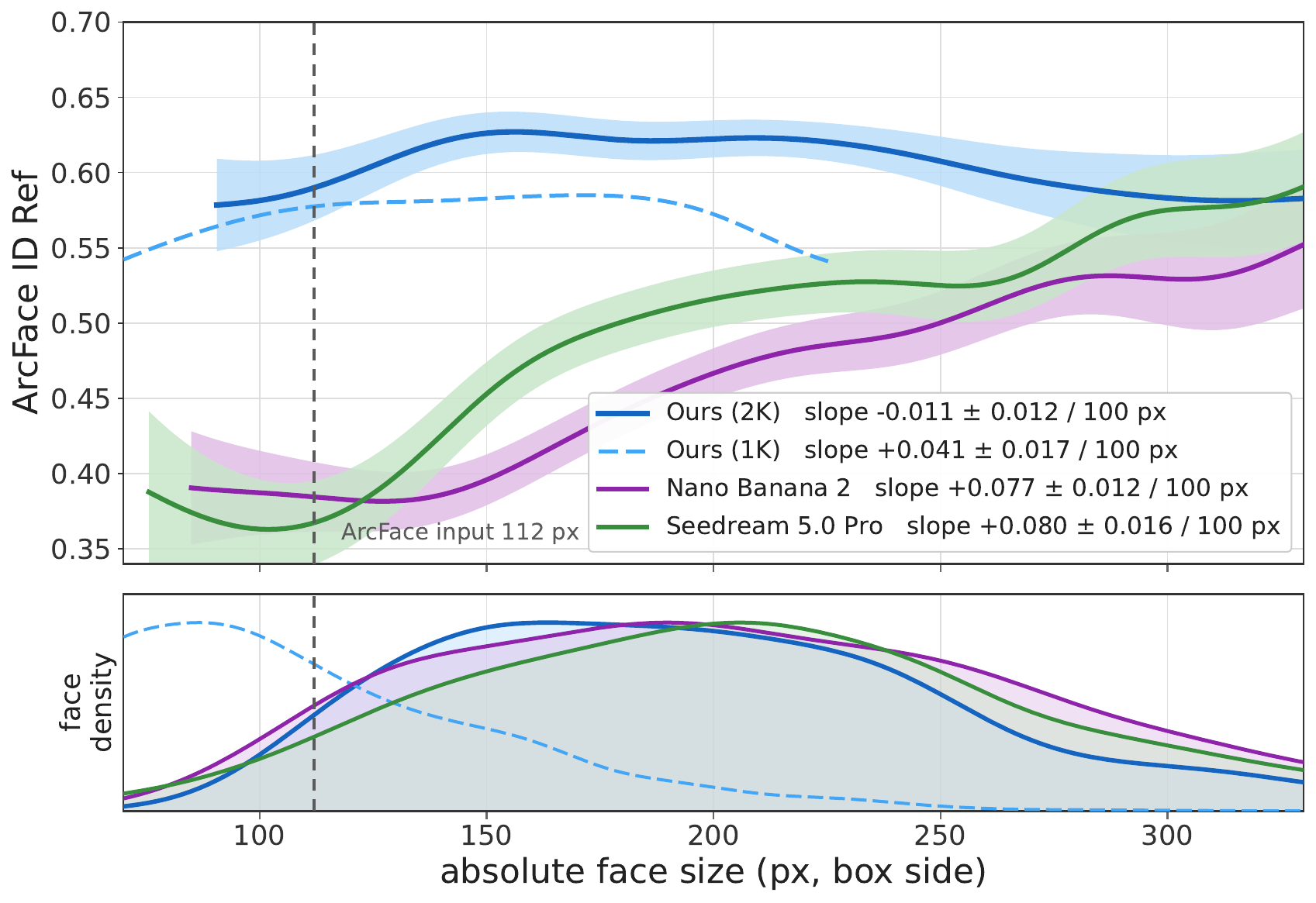}
  \caption{\small \textbf{Identity similarity against absolute face size.} Face size is measured in pixels. Curves are local means with 95\% confidence bands, annotated with the fitted slope. The dashed line marks 112~px, the input resolution of the ArcFace encoder, below which a face has to be upsampled before comparison. The lower panel shows the face-size distribution of each system.}
  \label{fig:face-size-abs}
\end{figure}

This section expands the analysis summarized in Section~\ref{sec:experiment-analysis} and refers to Figures~\ref{fig:face-size} and~\ref{fig:face-size-abs}. For every generated face on the benchmark we record its ArcFace similarity to the reference it is bound to, together with the size of its box, measured both as a fraction of the image side and in pixels. The analysis is per face, and every statement below concerns conditional means. The per-face standard deviation is 0.21--0.24 and is dominated by identity difficulty, pose, and occlusion, so face size accounts for only a small share of the variance of any single face.

Every baseline depends strongly on relative face size, as Figure~\ref{fig:face-size-rel} shows. Nano Banana 2 gains $0.016\pm0.003$ of similarity per percentage point of relative face side, Seedream 5.0 Pro $0.017\pm0.003$, and GPT-Image-2 $0.019$ $([+0.015,+0.023])$. \methodname{} at 2K is flat, at $-0.002\pm0.002$. This matters because growing the group barely changes how large the faces are: going from five to ten references moves the mean relative face side only from 11.9\% to 9.4\%. Across that narrow interval, shown in grey in the figure, Nano Banana 2 loses 0.040 of similarity and Seedream 5.0 Pro loses 0.042, while \methodname{} gains 0.006.

We then fit similarity on the number of references across all six group sizes, once with and once without face size as a covariate. For GPT-Image-2 and Seedream 5.0 Pro, controlling for face size removes most of the group-size slope, and what remains has an interval that includes zero; shrinking faces are therefore associated with a substantial part of their degradation. For \methodname{} the slope does not move, staying at $-0.014$ with an interval that excludes zero. The reduction in face resolution thus does not explain our remaining degradation, but this analysis does not distinguish cross-identity interference from occlusion, compositional complexity, or other factors that vary with group size. We do not quote a numerical share of the face-size association, because the interval on that ratio is too wide to be informative.

Figure~\ref{fig:face-size-abs} repeats the analysis in absolute pixels. The systems differ in output resolution here, so the curves have to be compared within a pixel band rather than along the axis. Within a common band, \methodname{} is the strongest below 200~px, and its margin is largest on the smallest faces: between 112 and 150~px it reaches 0.611, against 0.561 for GPT-Image-2, 0.388 for Seedream 5.0 Pro, and 0.383 for Nano Banana 2. GPT-Image-2 overtakes \methodname{} only in the 200--260~px band, at 0.654 against 0.616, and only 8\% of its faces fall there. Our advantage is therefore concentrated on small faces. This pattern is qualitatively consistent with GPT-Image-2's higher Copy-Paste, although it does not establish copying as the cause of the difference.

The 1K curve is a reference point rather than a controlled ablation, since it differs from the 2K curve in the reference-image configuration as well as in resolution. It is nonetheless informative about resolution. At 1K, 61\% of the faces are smaller than the 112~px input of the ArcFace encoder, against 3\% at 2K, so much of the benefit of high-resolution training reported in Appendix~\ref{sec:appendix-high-resolution} amounts to moving faces out of that regime.

\subsection{High-Resolution Training}
\label{sec:appendix-high-resolution}

\begin{figure*}[t]
  \centering
  \begin{subfigure}[t]{0.48\textwidth}
    \centering
    \includegraphics[width=\linewidth]{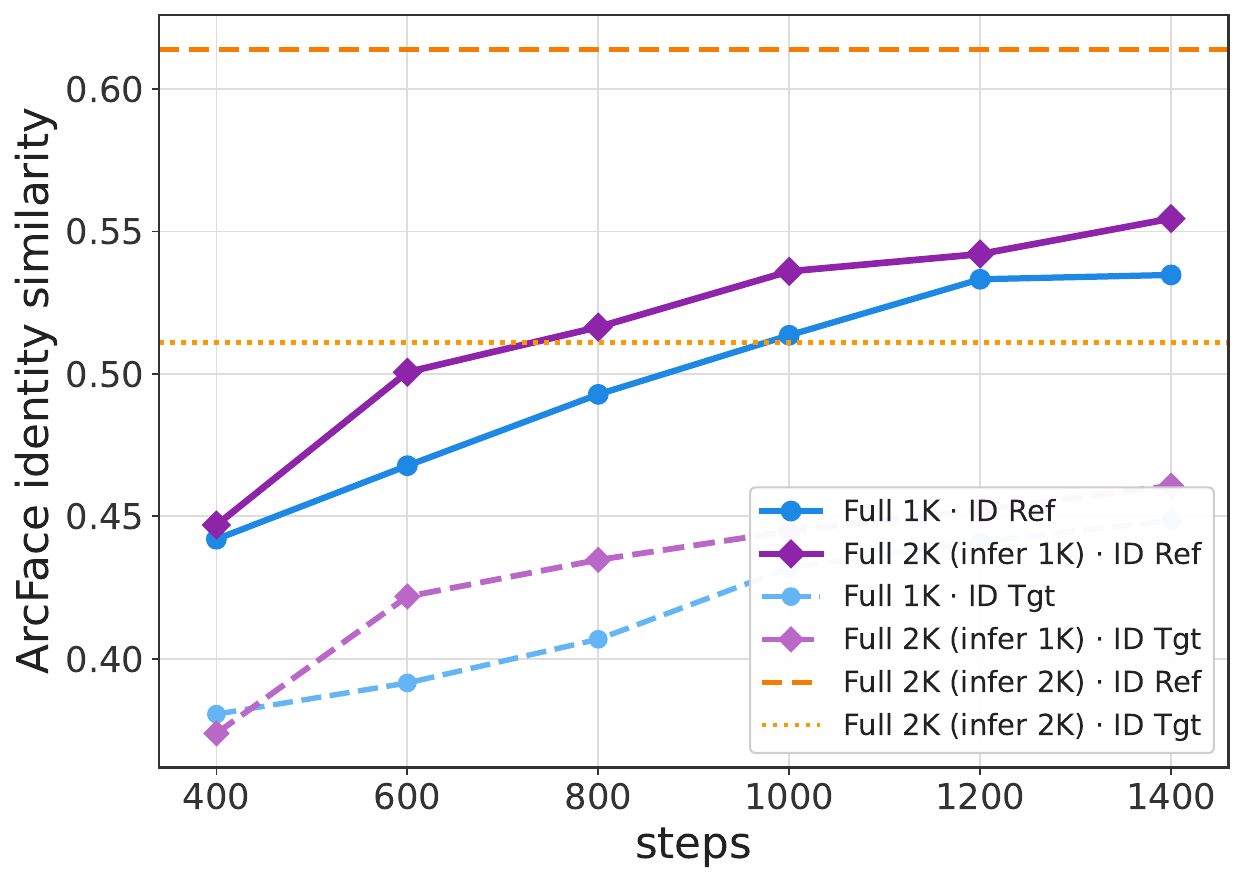}
    \caption{Identity similarity during training.}
    \label{fig:appendix-highres-identity}
  \end{subfigure}
  \hfill
  \begin{subfigure}[t]{0.48\textwidth}
    \centering
    \includegraphics[width=\linewidth]{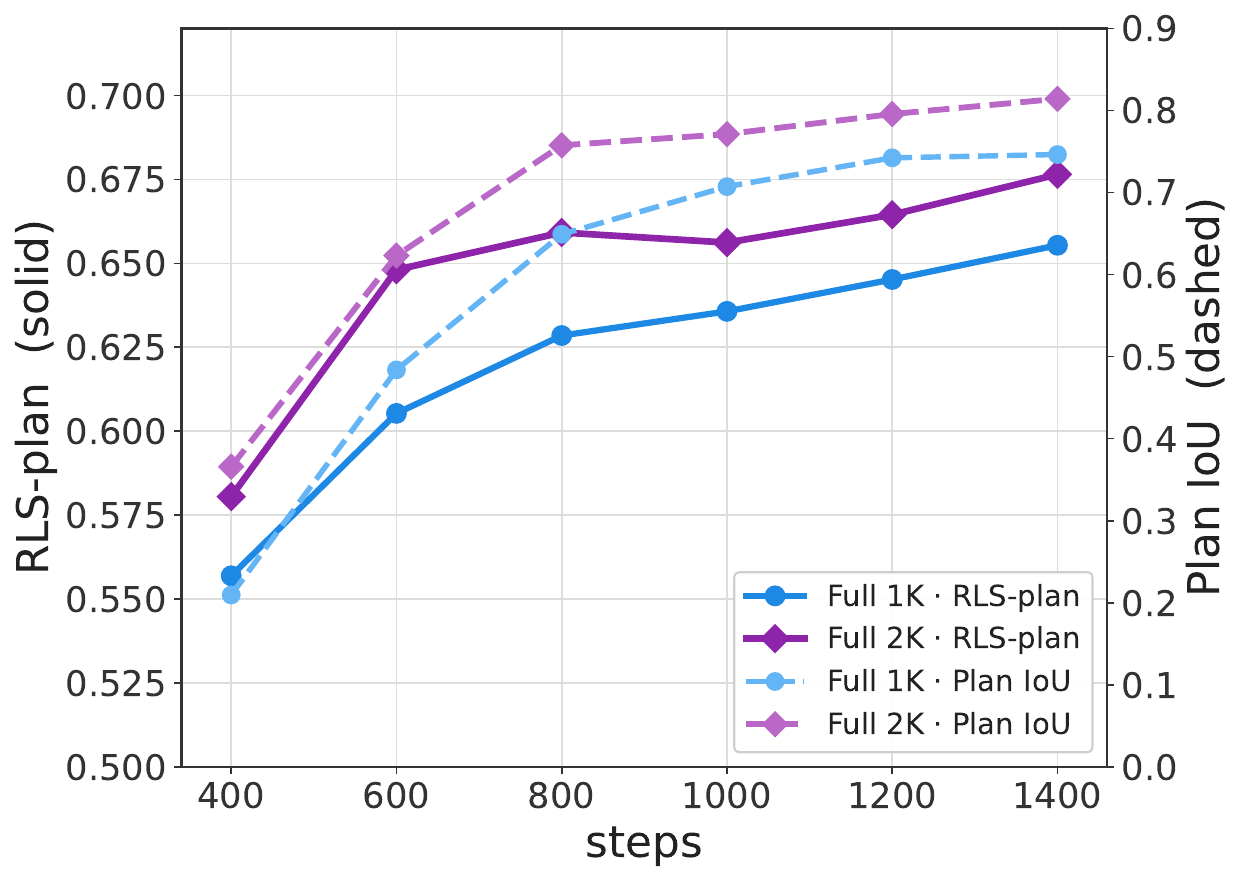}
    \caption{Planning quality and plan execution.}
    \label{fig:appendix-highres-planning}
  \end{subfigure}
  \caption{\small \textbf{High-resolution training.} We compare the Full model trained at 1K and 2K while evaluating both at 1K. The horizontal references in (a) show the final 2K benchmark scores from a different evaluation pipeline.}
  \label{fig:appendix-highres}
\end{figure*}

\begin{table*}[t]
  \centering
  \caption{\small \textbf{High-resolution comparison.} The first two rows use the same internal pipeline; the last row uses the main 2K benchmark pipeline.}
  \label{tab:appendix-highres}
  \small
  \begin{tabular}{lcccc}
    \toprule
    Train / test & Sim(Ref) $\uparrow$ & Sim(Tgt) $\uparrow$ & Layout$^{\ast}$ $\uparrow$ & Plan IoU $\uparrow$ \\
    \midrule
    1K / 1K & 0.546 & 0.460 & 0.740 & 0.773 \\
    2K / 1K & \textbf{0.555} & \textbf{0.461} & \textbf{0.759} & \textbf{0.814} \\
    2K / 2K$^\dagger$ & 0.614 & 0.511 & -- & -- \\
    \bottomrule
  \end{tabular}

  \vspace{0.2em}
  {\scriptsize $^{\ast}$Layout Score, defined in Appendix~\ref{sec:appendix-layout-score}. $^\dagger$Main benchmark pipeline; included as a contextual reference only.}
\end{table*}

Faces in a group image often occupy only a small fraction of the image. Increasing the target resolution therefore allocates more pixels to identity-critical facial regions and may also provide a denser spatial signal for executing the predicted layout. We study this effect by comparing the Full model trained at 1K with the Full model trained at 2K. For a controlled comparison, both models are first evaluated at 1K. We additionally report the final 2K benchmark result to show the operating point of the earlier checkpoint.

Under the controlled 1K evaluation, 2K training improves Sim(Ref) from 0.546 to 0.555 and does not reduce Sim(Tgt) or Layout Score. Its largest gain is in Plan IoU, which rises from 0.773 to 0.814. Figure~\ref{fig:appendix-highres-planning} further shows that the 2K model reaches stronger planning and execution scores throughout the later stage of training. These results indicate that high-resolution training mainly improves how faithfully image generation realizes the spatial plan, while providing a smaller identity benefit at a fixed inference resolution. When the 2K model is evaluated at its native resolution in the main benchmark, it reaches 0.614 Sim(Ref) and 0.511 Sim(Tgt); because that benchmark uses a different scoring pipeline, these values should not be interpreted as a controlled estimate of the gain from changing inference resolution alone.

\subsection{Representation Forcing Convergence}
\label{sec:appendix-rf-convergence}

\begin{figure*}[t]
  \centering
  \begin{subfigure}[t]{0.48\textwidth}
    \centering
    \includegraphics[width=\linewidth]{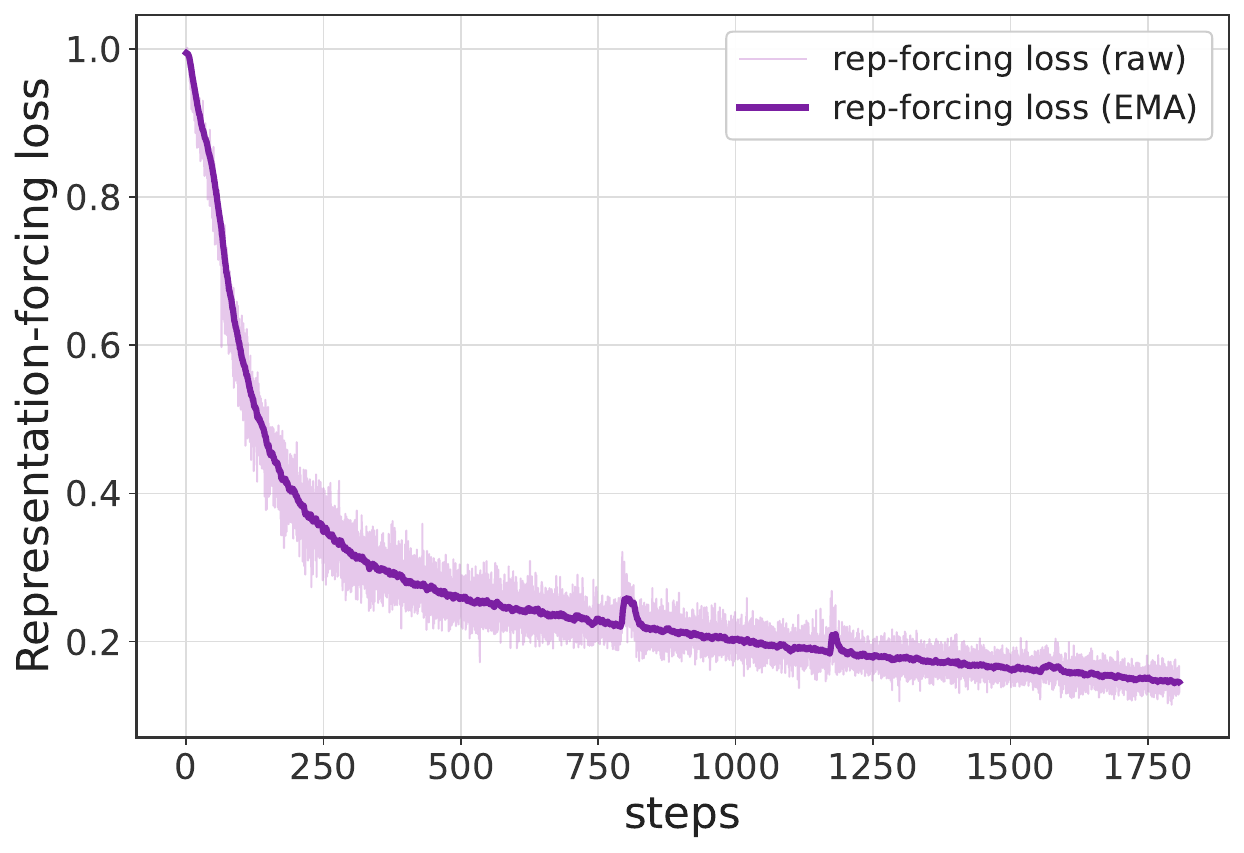}
    \caption{Representation Forcing training loss.}
    \label{fig:appendix-rf-loss}
  \end{subfigure}
  \hfill
  \begin{subfigure}[t]{0.48\textwidth}
    \centering
    \includegraphics[width=\linewidth]{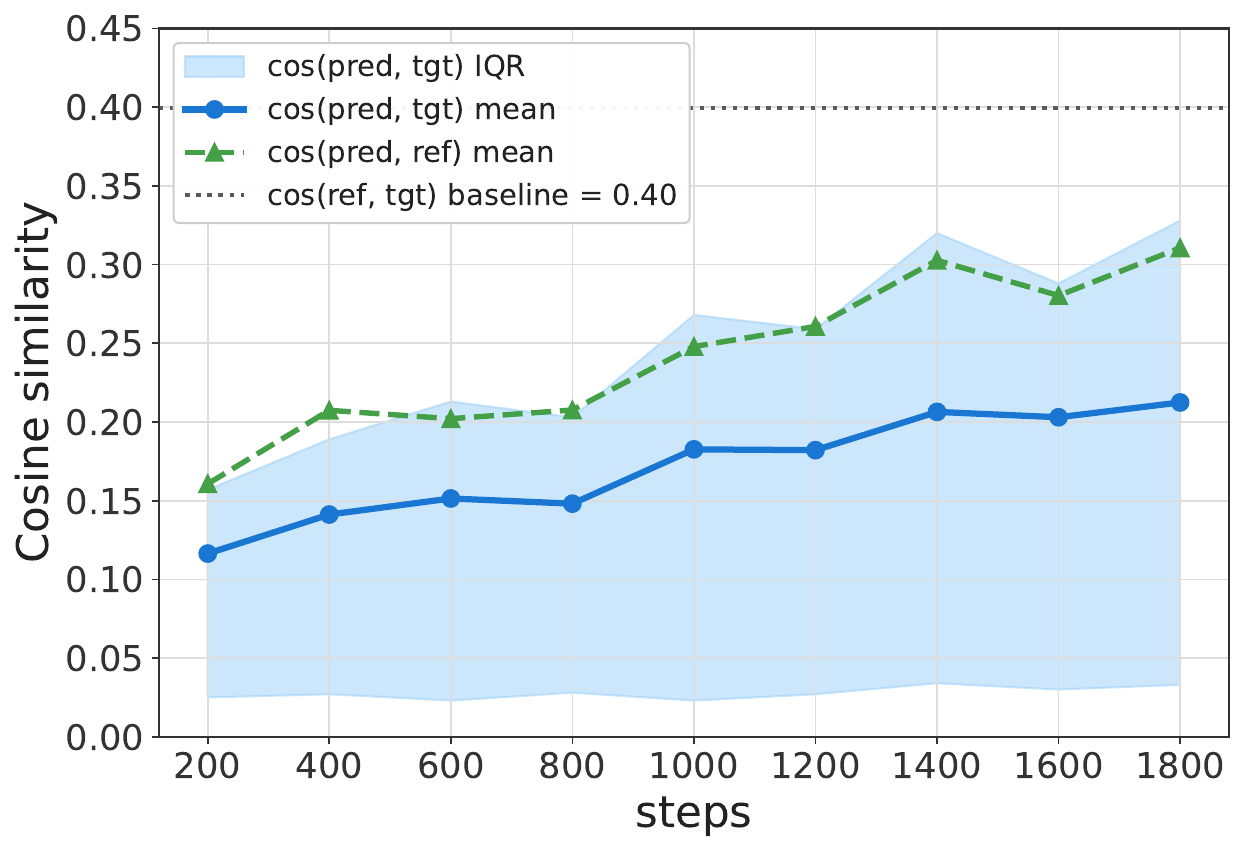}
    \caption{Inference-time representation alignment.}
    \label{fig:appendix-rf-cos}
  \end{subfigure}
  \caption{\small \textbf{Representation Forcing convergence.} The supervised identity prediction is steadily optimized during training and becomes increasingly aligned with both the target and reference identity embeddings.}
  \label{fig:appendix-rf-convergence}
\end{figure*}

Representation Forcing is designed to make the shared context retrieve identity information before image generation. Figure~\ref{fig:appendix-rf-loss} shows that its training objective can be optimized smoothly: the loss decreases from approximately 1.0 to 0.15. 

We further evaluate the predicted representation directly at inference time. As shown in Figure~\ref{fig:appendix-rf-cos}, the mean cosine similarity between the prediction and target identity embedding increases from 0.116 at 200 steps to 0.212 at 1800 steps. Its similarity to the reference identity follows the same trend. Both remain below the reference--target baseline of approximately 0.40, so the prediction does not fully reconstruct the target identity embedding. The result supports the narrower conclusion that Representation Forcing progressively aligns the prediction with the intended identity direction, consistent with improved identity addressability.

\subsection{LG-ID Loss: Algorithm and Hyper-parameters}
\label{sec:appendix-idloss}

\subsubsection{One-Step Identity Supervision}
\label{sec:appendix-idloss-algorithm}

\begin{figure*}[t]
  \centering
  \includegraphics[width=0.96\textwidth]{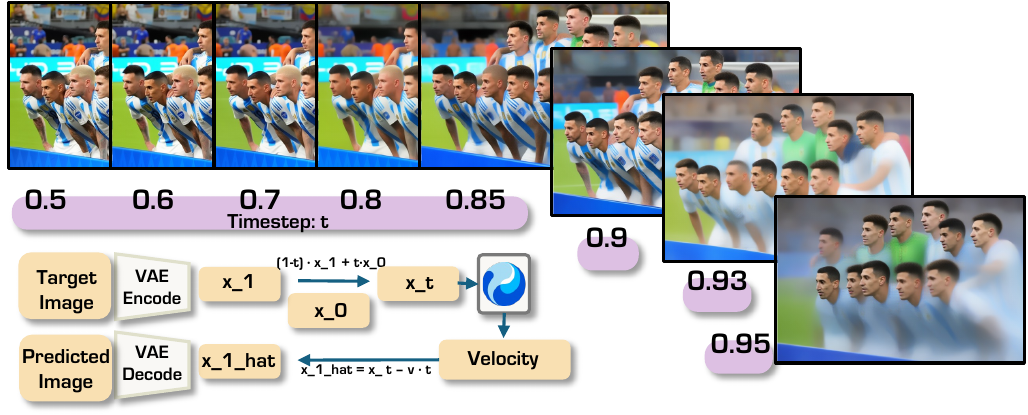}
  \caption{\small \textbf{Training-time one-step prediction and its outputs under different sampled timesteps.} We use $t=0.85$ as the timestep threshold and compute the LG-ID Loss only when $t \leq 0.85$.}
  \label{fig:appendix-idloss-one-step}
\end{figure*}

The LG-ID Loss augments the image flow-matching objective with identity supervision on a differentiable clean-image estimate. Given a noisy latent $\mathbf{x}_t$ and predicted velocity $\mathbf{v}_{\theta}(\mathbf{x}_t,t)$, we first compute
\begin{equation}
  \hat{\mathbf{x}}_{\mathrm{clean}} = \mathbf{x}_t - t\,\mathbf{v}_{\theta}(\mathbf{x}_t,t),
\end{equation}
decode $\hat{\mathbf{x}}_{\mathrm{clean}}$ with the VAE, and compare the resulting faces with their target identities using a frozen ArcFace encoder. Figure~\ref{fig:appendix-idloss-one-step} summarizes the one-step prediction and visualizes its output at different sampled timesteps.

For multi-person supervision, we avoid reordering identities with an embedding-based matching heuristic. The correspondence is already known from the annotation that supervises the Layout CoT: each referenced identity carries a fixed target face bounding box and its landmarks, so cropping the predicted and target images at these regions makes every crop pair correct by construction, and each identity is supervised on its own, independently of how many people share the image. We detect faces in the ground-truth image and match each detection to its dataset bounding box by IoU; bystanders without a referenced identity are ignored. The same ground-truth landmarks then determine corresponding crops in the predicted and target images. The LG-ID Loss is the mean per-face cosine distance, $1-\cos(\mathbf{e}^{\mathrm{pred}},\mathbf{e}^{\mathrm{tgt}})$, over valid referenced identities.

\subsubsection{Timestep Threshold}
\label{sec:appendix-idloss-threshold}

At high-noise timesteps, the one-step estimate may not contain a recognizable face, making identity supervision unreliable. As shown in Figure~\ref{fig:appendix-idloss-one-step}, the estimated clean image retains clear identity cues up to $t=0.85$, whereas facial details degrade rapidly at larger timesteps. We therefore set the hard timestep threshold to $0.85$ and compute the LG-ID Loss only when $t \leq 0.85$.

\subsubsection{Loss Weight}
\label{sec:appendix-idloss-weight}

\begin{figure*}[t]
  \centering
  \begin{subfigure}[t]{0.32\textwidth}
    \centering
    \includegraphics[width=\linewidth]{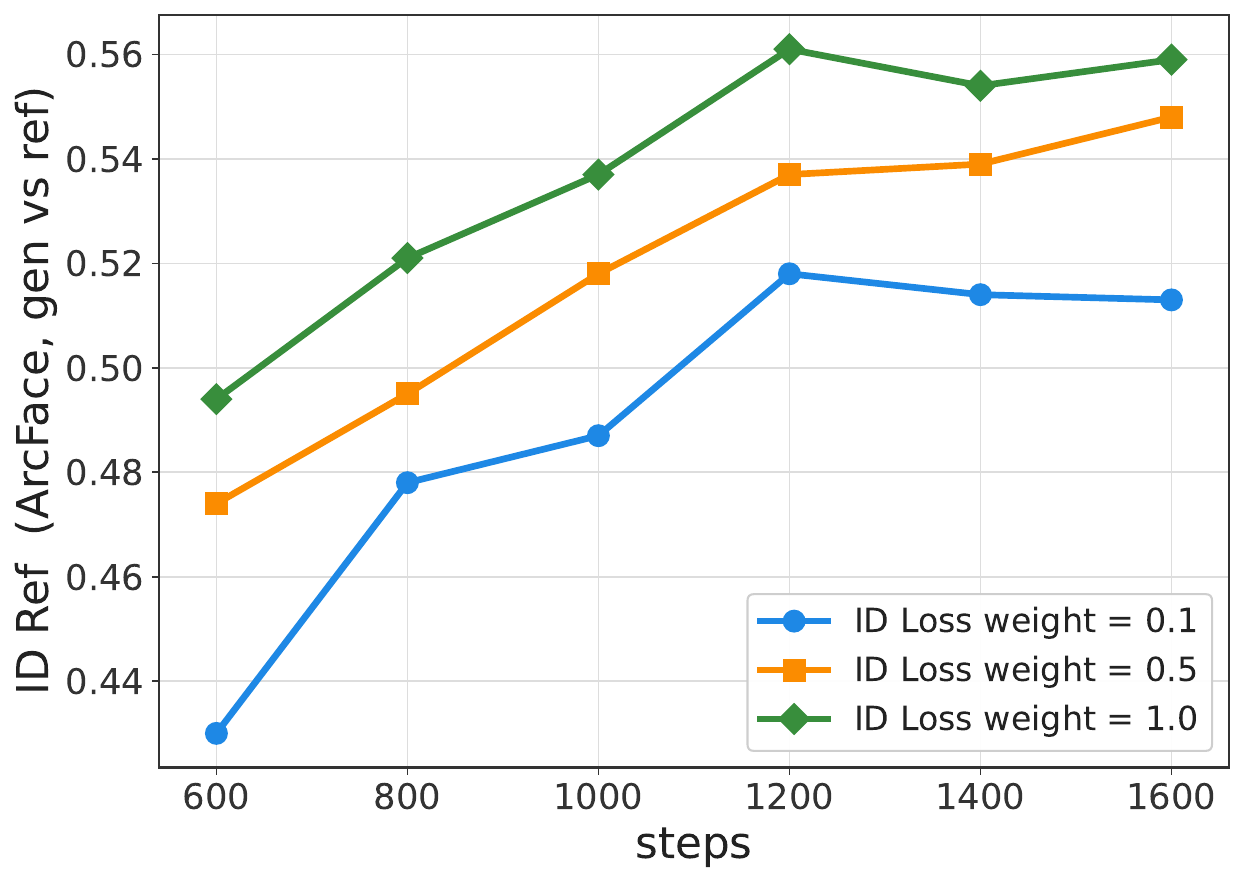}
    \caption{Identity preservation.}
    \label{fig:appendix-idloss-weight-id}
  \end{subfigure}
  \hfill
  \begin{subfigure}[t]{0.32\textwidth}
    \centering
    \includegraphics[width=\linewidth]{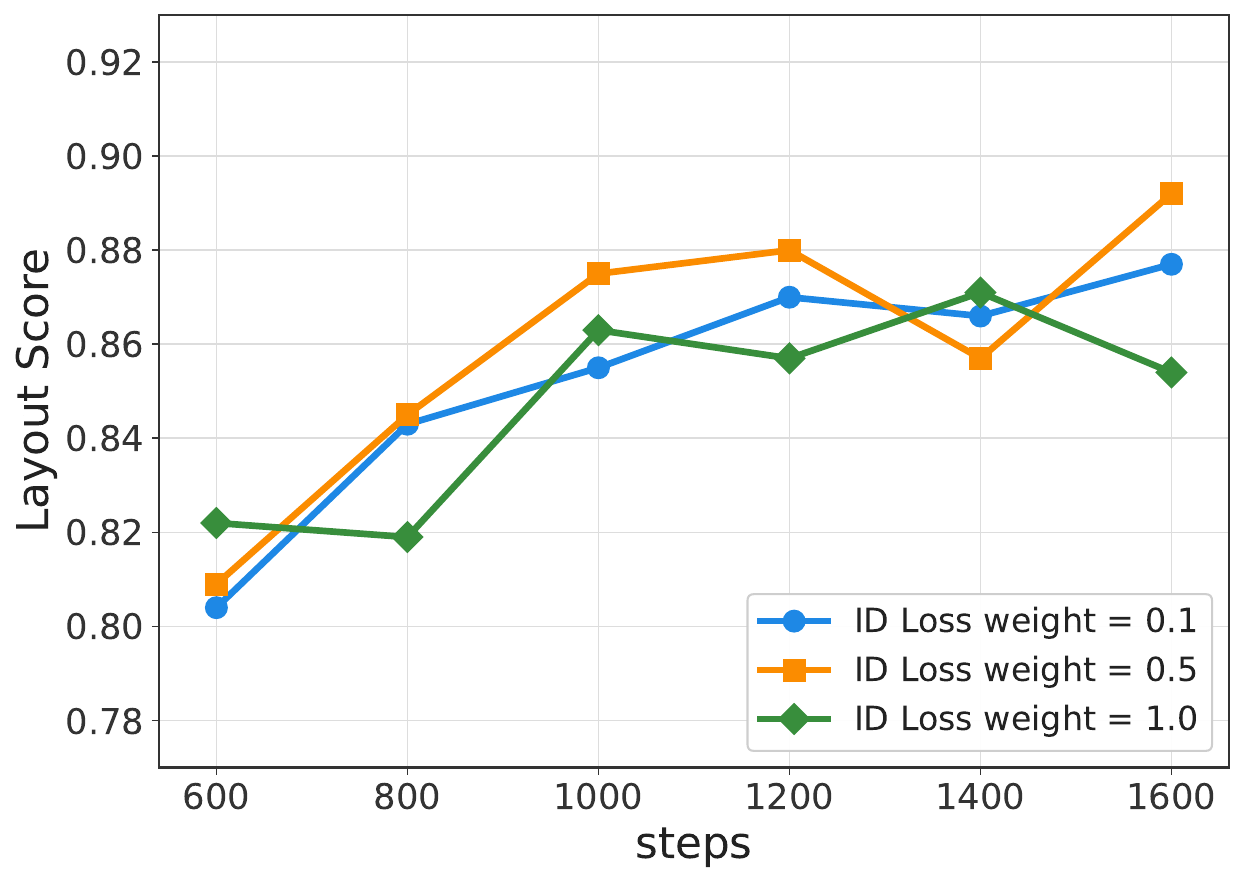}
    \caption{Layout Score.}
    \label{fig:appendix-idloss-weight-layout}
  \end{subfigure}
  \hfill
  \begin{subfigure}[t]{0.32\textwidth}
    \centering
    \includegraphics[width=\linewidth]{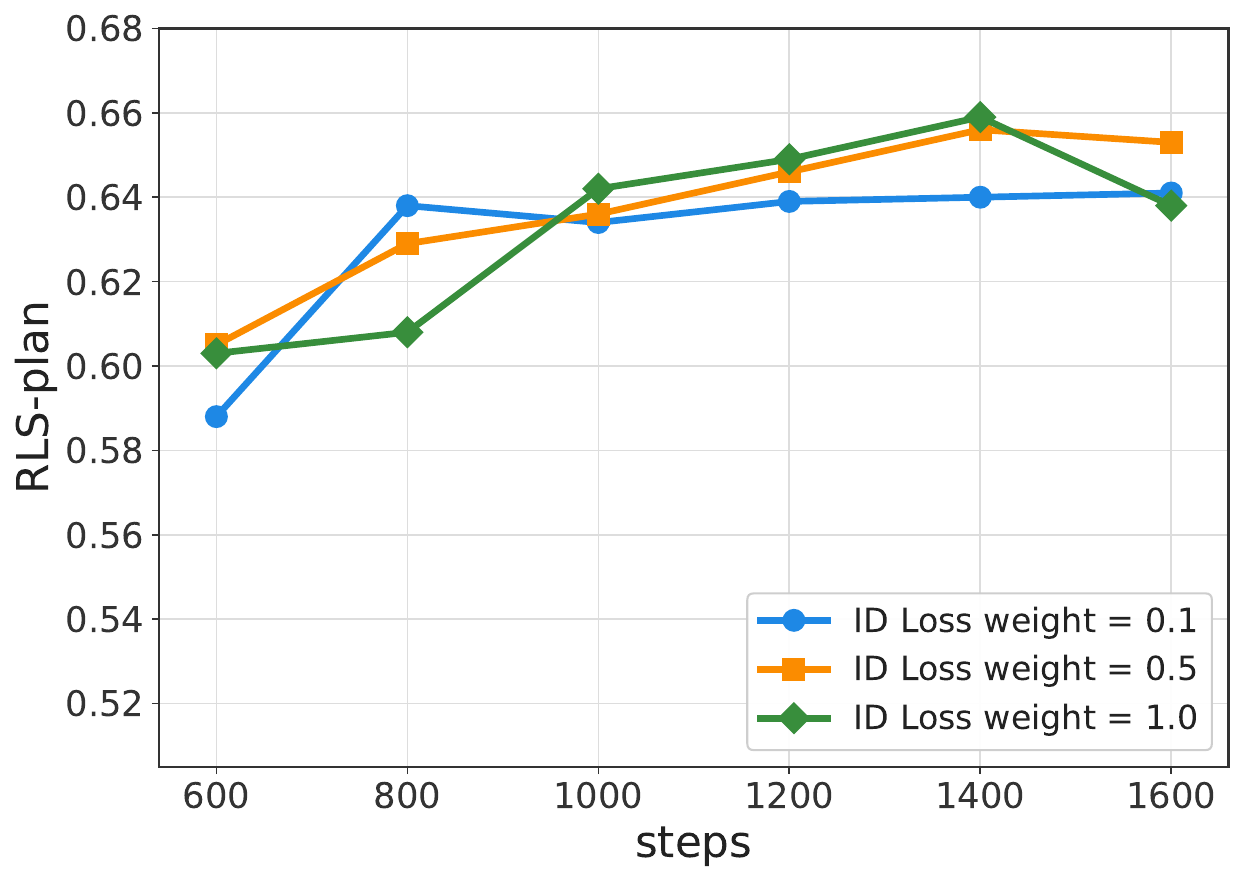}
    \caption{RLS.}
    \label{fig:appendix-idloss-weight-rls}
  \end{subfigure}
  \caption{\small \textbf{LG-ID Loss weight ablation.} Increasing $\lambda_{\mathrm{ID}}$ substantially improves Sim(Ref), whereas Layout Score and RLS remain in the same range without a stable ordering across weights.}
  \label{fig:appendix-idloss-weight}
\end{figure*}

We vary only the LG-ID Loss weight while keeping the remaining Full-model recipe fixed. As shown in Figure~\ref{fig:appendix-idloss-weight-id}, a larger weight consistently improves identity preservation at both training steps: Sim(Ref) at 1200 steps is 0.518, 0.537, and 0.561 for $\lambda_{\mathrm{ID}}=0.1$, $0.5$, and $1.0$, respectively; at 1600 steps, the corresponding values are 0.513, 0.548, and 0.559. The main experiments use $\lambda_{\mathrm{ID}}=0.5$, and this ablation shows that identity preservation can be pushed further by raising the weight.

The gain in identity does not produce a corresponding degradation in planning. Figures~\ref{fig:appendix-idloss-weight-layout} and~\ref{fig:appendix-idloss-weight-rls} show tightly overlapping trajectories across the three weights for the two composite layout measures, the Layout Score of Appendix~\ref{sec:appendix-layout-score} and the Relative Layout Score (RLS) of Appendix~\ref{sec:appendix-rls}: Layout Score remains within approximately 0.80--0.89, and RLS within approximately 0.59--0.66. Larger identity weights can slow early layout improvement slightly, but the difference disappears later in training. Within the tested range, increasing $\lambda_{\mathrm{ID}}$ therefore primarily affects identity preservation and has no visible cost in converged layout quality.

\subsection{Effect of the LG-ID Loss}
\label{sec:appendix-idloss-effect}

Table~\ref{tab:appendix-idloss-effect} expands the main-paper ablation of the LG-ID Loss with the complete set of identity and layout measures, including the Layout Score defined in Appendix~\ref{sec:appendix-layout-score}. The loss alone raises Sim(Ref) from 0.339 to 0.506 and Sim(Tgt) from 0.304 to 0.435. At the same time, Layout Score increases slightly from 0.690 to 0.700, Count from 0.771 to 0.845, and Coverage from 0.741 to 0.947. Direct identity supervision therefore does not trade layout validity for identity similarity. Because each identity is supervised inside its annotated region, the loss is reduced by putting the right person in the right place, so it carries a spatial signal in addition to an identity one.

\begin{table}[H]
  \centering
  \caption{\small \textbf{Effect of the LG-ID Loss.} It gives the largest individual identity gain while preserving or improving all reported layout measures.}
  \label{tab:appendix-idloss-effect}
  \resizebox{\columnwidth}{!}{
  \begin{tabular}{lccccc}
    \toprule
    Variant & Sim(Ref) $\uparrow$ & Sim(Tgt) $\uparrow$ & Layout $\uparrow$ & Count $\uparrow$ & Coverage $\uparrow$ \\
    \midrule
    Default & 0.339 & 0.304 & 0.690 & 0.771 & 0.741 \\
    + ID Loss & 0.506 & 0.435 & 0.700 & 0.845 & 0.947 \\
    Full & \textbf{0.555} & \textbf{0.461} & \textbf{0.759} & \textbf{0.869} & \textbf{0.960} \\
    \bottomrule
  \end{tabular}}
\end{table}

The complete configuration combines the LG-ID Loss with ID tokens, Representation Forcing, Layout CoT, and additional text-to-layout supervision, raising Sim(Ref) and Sim(Tgt) to 0.555 and 0.461, respectively, while reaching a Layout Score of 0.759. This result improves over the LG-ID-only branch, but because several components and the training corpus change together, the comparison does not isolate the source of the additional gain.

\subsection{Setup and Reading of the Ablation Branches}
\label{sec:appendix-ablation-details}

All variants in Table~\ref{tab:ablations} start from the same base model and are trained independently under the same recipe, with sample-wise rather than fixed training resolutions, and are evaluated at 1K with ArcFace-only identity scoring. P1 is the default model, trained on the same dataset but without Layout CoT or ID tokens. P2 and P3 are both trained with Layout CoT; at inference, P2 uses the layout the model predicts itself, whereas P3 replaces it with the ground-truth layout. Because the P3 layout is derived from the target image, P3 is an oracle upper bound rather than an estimate of an achievable planning gain. It establishes that high-quality layout conditions can improve both composition and identity preservation, while the gap between P2 and P3 identifies planning quality as an important remaining source of error.

P7 additionally trains on the text-to-layout corpus of Appendix~\ref{sec:appendix-data} and scores higher than P2 on both identity, with Sim(Ref) of 0.555 versus 0.364 and Sim(Tgt) of 0.461 versus 0.316, and composition, with Count of 0.869 versus 0.828 and Coverage of 0.960 versus 0.813. It also adds ID tokens, Representation Forcing, and the LG-ID Loss, however, so this comparison does not isolate the contribution of the corpus.

P4 adds an ID token to P1, and P5 adds Representation Forcing on top of P4. The ID token alone yields small but consistent gains, the largest of which is Count, from 0.771 to 0.827. Representation Forcing then raises Sim(Ref) from 0.351 to 0.364 and Sim(Tgt) from 0.313 to 0.328. In the three examples shown in Figure~\ref{fig:rf-attention}, each supervised identity prediction preferentially attends to its corresponding ID token. The end-to-end gains are modest, so we interpret this evidence as consistent with improved identity addressability rather than as a large direct source of identity gain, and the attention analysis does not establish the effect relative to a model without Representation Forcing, nor equally strong routing from generated face patches to the ID tokens.

\section{Experimental Details}
\label{sec:appendix-experimental-details}

\subsection{Training Sequence and Data Construction}
\label{sec:appendix-data}
The causal training sequence is organized as follows: \textbf{1)} reference images and the prompt; \textbf{2)} reference-identity selection and ID loading; \textbf{3)} face layout and identity binding, followed by body-region and pose planning; \textbf{4)} rendering the plan into a visual layout condition; \textbf{5)} a summary and detailed recaption; \textbf{6)} target identity representation prediction; and \textbf{7)} target image generation. Stage 5 restates the request after the plan has been committed, giving the generator a next-token-supervised language description that is already consistent with the rendered canvas instead of only the short user prompt. This order is the one used in training and inference; Figure~\ref{fig:pipeline} arranges some stages differently for visual layout only, and the order stated here takes precedence.

Each group-image training sample contains a target image, reference images, and reference-to-target identity correspondences, with short captions as user prompts and detailed captions as recaptions. Face detection and ArcFace provide reference and target identity embeddings, while person detection and pose estimation by YOLOv11 \citep{yolo11_ultralytics} provide person regions and keypoints. Together with the captions and correspondences, these signals supervise the structured reasoning stages and let the renderer produce the layout condition. We additionally use text-to-layout examples to supervise layout reasoning without a target image, strengthening planning independently of full image-generation samples.

\subsection{Identity Loading and Representation Forcing}
\label{sec:appendix-id-details}
The prompt may require only a subset of the available reference people and may also describe people without a reference image, so the model first identifies the participating identities. Selection and binding are ordered rather than mutually dependent: the model emits the list of selected identities and loads one ID token per selection, and the binding of each token to a concrete planned person follows in the layout stage. This identity--layout binding gives each loaded identity a spatial role rather than treating the references as an unordered collection. The ID embeddings complement the low-level reference-image features with high-level identity-specific information, analogous to the complementary roles of ViT \citep{dosovitskiy2020image} and VAE embeddings at different feature levels: for each selected person, a 512-dimensional ArcFace embedding is mapped to the model hidden dimension by a lightweight MLP and injected at the corresponding ID-token position.

Loading ID tokens does not by itself guarantee that image generation will use them. Each person is represented by a single continuous token, projected from one 512-dimensional vector, inside an interleaved sequence of 10--20K tokens, so generation-side attention over the injected identity tokens can be diffuse and identity-agnostic. For every selected reference identity with a target correspondence, Representation Forcing therefore places a representation token before the target image. The backbone computes its hidden state from the preceding prompt, visual references, ID tokens, and layout reasoning, and the output projector $g_{\mathrm{out}}$ maps it to the ArcFace space, where it is compared with the ArcFace embedding of the corresponding person in the target image. The resulting hidden states remain causally available to the later image tokens, forming an identity scaffold for generation.

\subsection{Training Configuration}
\label{sec:appendix-training-config}
\methodname{} uses a mixture-of-transformers backbone with 60B parameters on the understanding side and 60B parameters on the generation side, initialized from an in-house base model. Training uses 400K in-house group-image samples; Figure~\ref{fig:data-pipeline} reports the identity-occurrence and reference-count statistics of this corpus. We optimize with Muon \citep{liu2025muon} at a learning rate of $1\times10^{-5}$ on the generation side and $3\times10^{-6}$ on the understanding side, with a packed sequence length of 72K tokens. Training runs on 128 H20 GPUs, and all reported models are taken from 1600 iterations. The objective weights are $\lambda_{\mathrm{FM}}=1.0$, $\lambda_{\mathrm{RF}}=1.0$, and $\lambda_{\mathrm{ID}}=0.5$.

\subsection{Shared Evaluation Protocol}
Existing multi-identity benchmarks focus on a small number of people and therefore do not measure how identity preservation scales to larger groups, which is why we construct the five-to-ten-person benchmark of Section~\ref{sec:experimental-setup}. The benchmark contains 210 examples, but Table~\ref{tab:main-results} aggregates available valid scores separately for each model and metric. Ours has 210 generated samples and GPT-Image-2.5 has 199; undefined face metrics can further reduce effective counts. Missing generations are not assigned zero scores, and these aggregates are not a matched-cohort significance analysis. Each evaluation example contains one generated image, 5--10 reference identities, a target group image, and reference-to-target identity correspondences. Face detections and embeddings are computed independently for the generated, reference, and target images. For models with Layout CoT, we additionally parse face boxes, body boxes, pose keypoints, and identity--layout bindings from the generated plan. Body-level evaluation uses YOLO11x-pose with a confidence threshold of 0.25 and an input size of 640; if body detection is unavailable, only face-level terms are retained.

\subsection{Identity Metrics}
The main comparison follows the WithAnyone protocol and averages results from ArcFace, FaceNet, and AdaFace. For each backbone, generated faces are matched to reference or target faces by maximum-similarity Hungarian assignment. The resulting means define Sim(Ref) and Sim(Tgt), respectively. Copy-Paste, introduced and validated against human judgment by WithAnyone~\citep{xu2025withanyone}, measures the scale of copy-paste artifacts. It compares each matched generated--target--reference triplet in angular embedding space:
\begin{equation}
  \mathrm{Copy\text{-}Paste}
  =
  \frac{\theta_{gt}-\theta_{gr}}{\theta_{tr}},
\end{equation}
where $\theta_{ab}=\arccos(\cos(\mathbf{e}_a,\mathbf{e}_b))$. A larger value indicates that the generated face is closer to the reference than to its target-context appearance. The ablation Sim(Ref) and Sim(Tgt) metrics use the same generated-to-reference and generated-to-target semantics with a single ArcFace encoder.

\subsection{Image and Text Similarity Metrics}
CLIP-I and DINO-I are the cosine similarities between the generated and the target image in the CLIP \citep{radford2021learning} and DINOv2 \citep{oquab2023dinov2} image-embedding spaces, and CLIP-T is the cosine similarity between the generated image and the user prompt in the joint CLIP space. Aes is the aesthetic-predictor-v2.5 SigLIP score. HPSv3 is the official raw preference mean (mu), using English benchmark captions and official preprocessing, without clipping or normalization to 0--10. Both are higher-is-better. WithAnyone Aes uses 176 historically scored images; HPSv3 uses all 202 successful generations. These three similarity metrics describe overall scene and prompt agreement and are insensitive to which identity appears where, so we read them alongside the identity metrics above rather than on their own.

\subsection{Layout Score}
\label{sec:appendix-layout-score}
Layout Score measures the overall validity of the final generated image as a weighted aggregate of seven sub-scores in $[0,1]$. If a component cannot be computed, it is omitted and the remaining weights are renormalized.

\paragraph{Count.}
Let $N$ be the expected number of people in the target and $M$ the number of detected generated faces. We compute
\begin{equation}
  S_{\mathrm{count}}
  =
  1-\min\left(1,\frac{|M-N|}{\max(1,N)}\right),
\end{equation}
so both missing and extra people are penalized relative to the target group size.

\paragraph{Coverage, uniqueness, and distinctness.}
We construct the cosine-similarity matrix between reference and generated faces. Maximum-similarity Hungarian assignment determines one-to-one matches, and a reference is covered when its matched similarity is at least 0.20; $S_{\mathrm{cover}}$ is the fraction of covered references. For uniqueness, a reference is marked as over-claimed if more than one generated face has similarity at least 0.20 to it, and $S_{\mathrm{uniq}}$ is one minus the over-claimed-reference rate. Table~\ref{tab:main-results} reports Dup as the forced-collision rate, not the over-claimed-reference rate. Distinctness captures the opposite collision: each reference independently selects its most similar generated face, and $S_{\mathrm{distinct}}$ is one minus the fraction of references whose best face is also selected by another identity. The Dup column of Table~\ref{tab:main-results} reports this distinctness collision rate, $1-S_{\mathrm{distinct}}$.

\paragraph{Sensitivity to the matching threshold.}
The 0.20 similarity threshold is permissive by design, so that a reference counts as covered whenever the generated person is recognizably the intended one. Coverage is nevertheless insensitive to it: for P7, raising the threshold to 0.25, 0.30, and 0.35 lowers Coverage by 0.010, 0.017, and 0.027, and lowering it to 0.15 raises Coverage by 0.012. Only 3.6\% of references have a matched similarity in $[0.15,0.30)$, against a median matched similarity of 0.554, so few decisions sit near the boundary.

\paragraph{Anonymous leakage and spatial validity.}
Generated faces that are not valid Hungarian matches are treated as anonymous. An anonymous face is counted as identity leakage if its similarity to any reference is at least 0.25; $S_{\mathrm{noleak}}$ is one minus the leaked-anonymous-face rate. Spatial validity is computed from pairwise overlaps. Face IoU above 0.30 and body IoU above 0.50 are treated as severe overlaps. When both detections are available, $S_{\mathrm{spatial}}$ is one minus the mean of the face- and body-overlap rates; otherwise, only the face-overlap rate is used.

\paragraph{Plan adherence.}
We match generated face boxes to the available layout boxes and average the matched IoUs. Ground-truth boxes are used when a ground-truth layout is available; otherwise, boxes parsed from the model's CoT plan are used. This mean IoU defines $S_{\mathrm{plan}}$.

\subsection{Relative Layout Score and Plan IoU}
\label{sec:appendix-rls}
The Relative Layout Score (RLS), adopted from ``R-spatial'' in T2I-R1~\citep{jiang2026t2i}, evaluates the relative configuration of the model's textual plan against the ground-truth layout, without using the generated image. It combines relative count, position, and size, all in $[0,1]$ and higher-is-better.

\paragraph{Person matching.}
Planned and ground-truth people with the same reference identity are matched directly. Remaining people are represented by face centers, or body centers when faces are unavailable. The centers in each layout are centered and normalized by their root-mean-square spread before Hungarian geometric matching, preventing global translation or scale from dominating correspondence.

\paragraph{Relative count.}
Face and body counts are scored independently with the same normalized cardinality rule as $S_{\mathrm{count}}$ above, using planned and ground-truth counts. The available face- and body-count scores are then averaged into $S^{\mathrm{rel}}_{\mathrm{count}}$.

\paragraph{Relative position.}
For every pair of matched people, we check whether their left--right and top--bottom ordering agrees between the plan and ground truth, giving the two order-consistency terms $S_{\mathrm{order}\text{-}x}$ and $S_{\mathrm{order}\text{-}y}$. We additionally align planned centers to ground-truth centers with a non-reflective Umeyama similarity transform and convert the normalized residual distances into an OKS-style soft alignment score $S_{\mathrm{align}}$. A fitted-scale penalty prevents a collapsed layout from receiving a high score merely because Procrustes alignment can enlarge it. The order-consistency and aligned-center scores jointly define the position term,
\begin{equation}
  S_{\mathrm{order}} = 0.6\,S_{\mathrm{order}\text{-}x} + 0.4\,S_{\mathrm{order}\text{-}y},
  \qquad
  S^{\mathrm{rel}}_{\mathrm{pos}} = 0.6\,S_{\mathrm{order}} + 0.4\,S_{\mathrm{align}},
\end{equation}
where the horizontal ordering receives the larger of the two order weights.

\paragraph{Relative and absolute size.}
For each matched face or body box, its area is divided by the mean matched-box area in the same layout. The clipped absolute log-ratio between planned and ground-truth normalized areas measures relative-size agreement. We also compare the mean absolute face area between the two layouts to retain a penalty when every planned person is uniformly too large or too small. The available face, body, and absolute-size scores are averaged into $S^{\mathrm{rel}}_{\mathrm{size}}$.

\paragraph{Aggregation.}
The three terms are combined as
\begin{equation}
  \mathrm{RLS}
  =
  0.20\,S^{\mathrm{rel}}_{\mathrm{count}}
  + 0.45\,S^{\mathrm{rel}}_{\mathrm{pos}}
  + 0.35\,S^{\mathrm{rel}}_{\mathrm{size}}.
\end{equation}
Because the position term is measured after similarity alignment and the size term is normalized within each layout, RLS is insensitive to a global translation, rotation, or scaling of the whole plan and instead scores the arrangement of people relative to one another. It does not penalize the person overlap that is normal in a group photograph, and it is not an absolute bounding-box IoU.

Plan IoU instead evaluates the generation side. Planned face boxes are matched to detected faces in the generated image, and the mean matched IoU is reported. When ground-truth layout is directly provided as the generation condition, the same computation uses ground-truth boxes. Thus, RLS measures whether the understanding side proposes an appropriate relative layout, whereas Plan IoU measures whether image generation executes the supplied plan, and Layout Score measures the validity of the final image, of which plan adherence is only one weighted sub-score.

\section{Extended Background}
\label{sec:appendix-background}

\paragraph{Generative backbones.}
Text-to-image generation was established by denoising diffusion models \citep{ho2020ddpm} built on convolutional U-Net denoisers \citep{ronneberger2015u} and moved into a latent space to make high-resolution synthesis tractable \citep{stablediffusion}. Later systems replaced the convolutional denoiser with transformer architectures \citep{vaswani2017attention,peebles2023scalable} and the discrete noise schedule with flow matching \citep{lipman2022flow}, the recipe followed by current open-weight models \citep{flux2024,fluxkrea,batifol2025flux}. Transformer backbones are permutation-equivariant over their token sequence up to positional encoding \citep{xu2024permutation}, which is what lets heterogeneous conditions such as reference images, ID tokens, and layout tokens be interleaved in a single sequence. Our generation side inherits this design, but its conditioning signals are produced by the understanding side of the same model rather than by an external encoder. Contrastively pre-trained image--text encoders \citep{radford2021learning,zhai2023siglip} remain the standard source of semantic image features and of the CLIP-based metrics reported in Section~\ref{sec:experiments}.

\paragraph{Reference-conditioned and identity-preserving generation.}
Before unified models, reference conditioning was usually attached to a frozen generator: adapters inject reference features through additional cross-attention \citep{ye2023ipadapter}, and side networks inject spatial conditions \citep{zhang2023adding}. Face-specific variants specialize this recipe to identity \citep{papantoniou2024arc2face,peng2024portraitbooth}, and attention routing extends it from one subject to several \citep{wang2024moa,guo2025musar}. A parallel line folds these tasks into a single model that generates and edits from interleaved inputs \citep{xiao2024omnigen,liu2025step1x,wu2025uno,wu2025uso}. Our setting differs in scale rather than in conditioning mechanism: with five to ten identities in one image, the limiting factor becomes how each identity is supervised and addressed, not how reference features enter the network.

\paragraph{Planning and representation objectives.}
Layout planning has also been delegated to a language model that converts the prompt into a layout for a separate diffusion renderer \citep{lian2023llm}, whereas we keep planning inside the same generative context so that the plan and the image share one state (Section~\ref{sec:cot}). Supervising an intermediate prediction against a target representation is a long-standing idea in representation learning \citep{oord2018representation}, which Section~\ref{sec:method} specializes to one supervised prediction per reference identity rather than a single global representation.

\section{Limitations, Future Work, and Responsible Use}
\label{sec:limitations}

\paragraph{Layout is hard to evaluate.}
We evaluate layout from two perspectives: whether the faces are correctly planned, and whether the plan follows the relative location relationships of the ground truth, for example whether the person on the left remains on the left. When the prompt is underspecified, however, which is common in real-world use, many layouts may be equally valid and layout quality becomes difficult to measure against a single reference. We leave this to future work.

\paragraph{Evaluation scope.}
Our conclusions are drawn from 210 examples on a single benchmark, and the groups with more references are the smallest, so per-group numbers should be read as a trend rather than as precise estimates. All identity measures inherit the behavior of the underlying face detectors and recognizers, whose accuracy is known to vary across demographic groups, and the compared systems differ in their maximum output resolution and in whether they can plan a layout on their own.

\paragraph{Data and responsible use.}
Training uses group images collected from publicly accessible sources solely for non-commercial research, as described in the Ethics statement. As with other identity-preserving generators, including the proprietary systems we compare against, identity-conditioned generation can be misused to produce images of real people without their consent, to place someone in a scene they never took part in, or to impersonate them. These risks are common to the task rather than specific to our method; deployments should require consent from the referenced people and pair generation with provenance signaling.

\clearpage
\subsection*{Ethics statement}
We constructed the training dataset from images openly posted on public websites, including news outlets, official social-media accounts, and other publicly accessible pages, and did not collect from private or access-restricted sources. To reduce potential harm, we excluded images of minors and intimate, medical, or otherwise sensitive situations. The dataset is intended solely for non-commercial academic research, and any access to the full dataset will be governed by a research-only license and accompanying usage guidelines. Identity-preserving generation, including that offered by the open and proprietary systems we compare against, can be misused to depict real people without their consent or to impersonate them; this risk is shared by the task in general rather than specific to our method. We recommend that any deployment obtain consent from the referenced people and disclose generated content through provenance signals such as watermarking. Appendix~\ref{sec:limitations} further discusses responsible use.

\subsection*{Reproducibility statement}
Sections~\ref{sec:method} and~\ref{sec:experiments} specify the model design, objectives, benchmark protocol, baselines, and metrics. Appendix~\ref{sec:appendix-experimental-details} documents data construction, training configuration, metric definitions, thresholds, and additional ablations. The in-house base model and training corpus limit exact end-to-end reproduction outside the authors' environment; the reported details are intended to support reimplementation and evaluation with comparable resources.

\end{document}